\documentclass[%
 aip,
 amsmath,amssymb,
 reprint,%
 floatfix,
]{revtex4-1}

\usepackage{graphicx}
\usepackage{graphics}
\usepackage{float}
\usepackage{dcolumn}
\usepackage{bm}
\usepackage[utf8]{inputenc}
\usepackage[T1]{fontenc}
\usepackage{mathptmx}
\usepackage{color,soul}
\usepackage{amssymb}
\usepackage{amsmath}
\usepackage{mathptmx} 
\usepackage{subfigure}
\usepackage{xcolor}
\usepackage{subfloat}
\usepackage{multirow}
\usepackage{rotating}
\usepackage{booktabs}
\usepackage{relsize}
\usepackage{nomencl}
\usepackage{etoolbox}
\usepackage{float}
\usepackage{url}
\usepackage{hyperref}
\usepackage[ruled]{algorithm2e}
\usepackage{nomencl}
\usepackage{xcolor}
\usepackage[normalem]{ulem}

\makenomenclature
\begin{document}

\newpage
\preprint{AIP/123-QED}

\title{Continual Learning of Range-Dependent Transmission Loss for Underwater Acoustic using Conditional Convolutional Neural Net}
\author{Indu Kant Deo}
\email{indukant@mail.ubc.ca}
\author{Akash Venkateshwaran}
\email{akashv22@student.ubc.ca}
\author{Rajeev K. Jaiman}
\email{rjaiman@mech.ubc.ca}

\affiliation{Department of Mechanical Engineering,  The University of British Columbia, Vancouver, BC V5T 1Z4, Canada}

\date{\today}

\begin{abstract}
There is a significant need for precise and reliable forecasting of the far-field noise emanating from shipping vessels.
Conventional full-order models based on the Navier-Stokes equations are unsuitable, and sophisticated model reduction methods may be ineffective for accurately predicting far-field noise in environments with seamounts and significant variations in bathymetry. 
Recent advances in reduced-order models, particularly those based on convolutional and recurrent neural networks, offer a faster and more accurate alternative. These models use convolutional neural networks to reduce data dimensions effectively.
However, current deep-learning models face challenges in predicting wave propagation over long periods and for remote locations, often relying on auto-regressive prediction and lacking far-field bathymetry information. This research aims to improve the accuracy of deep-learning models for predicting underwater radiated noise in far-field scenarios. We propose a novel range-conditional convolutional neural network that incorporates ocean bathymetry data into the input. By integrating this architecture into a continual learning framework, we aim to generalize the model for varying bathymetry worldwide. 
To demonstrate the effectiveness of our approach, we analyze our model on several test cases and a benchmark scenario involving far-field prediction over Dickin’s seamount in the Northeast Pacific. Our proposed architecture effectively captures transmission loss over a range-dependent, varying bathymetry profile. This architecture can be integrated into an adaptive management system for underwater radiated noise, providing real-time end-to-end mapping between near-field ship noise sources and received noise at the marine mammal's location.

\end{abstract}

\maketitle

\section{Introduction}

\begin{figure}[htbp!]
    \centering
    \includegraphics[width=0.99\linewidth]{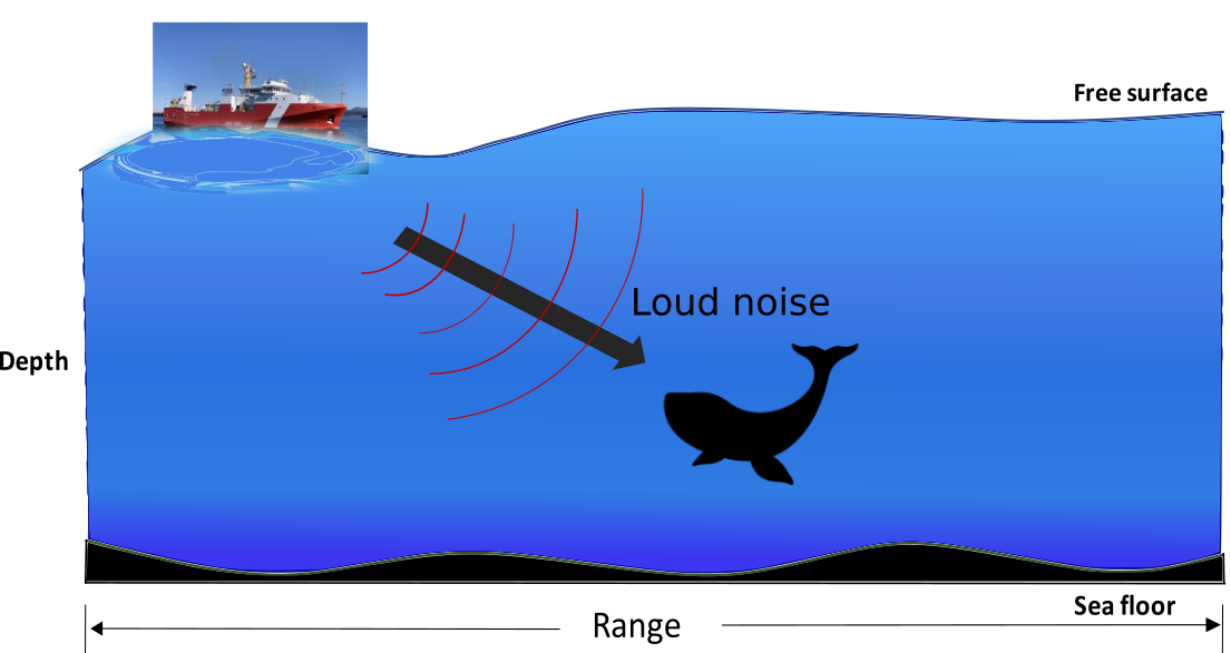}
    \caption{Illustration of real-time noise prediction from a marine vessel at a distant mammal location.}
\label{fig:real_time_ocean}
\end{figure}

Marine vessel activities are a major source of underwater ocean noise, which poses a significant threat to marine ecosystems \cite{duarte2021soundscape}. 
Solving this problem requires a collaborative, interdisciplinary approach involving environmentalists and engineers. Recent advancements in naval architecture and engineering offer promising ways to reduce noise through innovative green vessel design and operational strategies.
Intelligent Green Marine Vehicles (IGMVs) are crucial in addressing the issue of underwater ocean noise. These next-generation IGMVs need to be adaptable and continuously learn from their environment to adjust their operations. For example, one of the effective operational strategies is nearly real-time optimization of noise emissions from marine vessels, particularly in areas frequented by marine mammals \cite{venkateshwaran2024multi}.
Moreover, they must continually learn about the varied bathymetry environments encountered during operations.
This real-time prediction capability empowers vessels to adjust their operations dynamically to minimize acoustic disturbance to marine life. The illustration in Figure \ref{fig:real_time_ocean} depicts a scenario of ship operation integrating real-time noise prediction, displaying the significance of this capability in mitigating environmental impact.

Forecasting the underwater noise generated by marine vessels requires a thorough understanding of the dynamics of underwater radiated noise (URN).
Underwater ocean noise represents a complex wave propagation phenomenon, involving interactions between acoustic wavefronts and the ocean's surface and floor \cite{jensen2011computational}. 
The dynamics of underwater acoustic wave propagation are governed by hyperbolic partial differential equations (PDEs), which stem from conservation laws \cite{leveque2002finite}. Consequently, there has been increasing attention within the research community towards developing efficient and accurate computational solutions for these underlying PDEs.
Utilizing advanced numerical discretization techniques to solve these partial differential equations can result in significant computational overheads, making them impractical for tasks requiring real-time prediction. 
To create an adaptable marine system, it is crucial to have the capability to predict the far-field noise spectrum generated by vessels within complex and dynamic environments in nearly real-time.
Conventional wisdom dictates that a lower-dimensional reduced-order model, which can effectively capture the physics of the problem and produce solutions in nearly real-time, should replace the full-order PDE model.

Conventional approaches employ the wave equation to create simplified models for predicting sound propagation in the ocean, aiming to enhance computational efficiency. Examples of such methods include ray models \cite{porter1987gaussian, porter2019beam} and normal mode models \cite{williams1970normal, chapman1990normal}. These models require various environmental factors like sound speed, seabed properties, and bathymetry as input to predict accurate sound propagation. While physics-based acoustic solvers offer valuable insights, they pose challenges to integrate data from different sources and with varying levels of accuracy, such as simulation data and experimental measurements.
Consequently, it is essential to build a flexible data-driven model capable of assimilating ocean acoustics data from various sources and environmental conditions. 
Such a general-purpose model could be employed to evaluate underwater ocean acoustic propagation for varying bathymetry and different ocean environment conditions. 
Nevertheless, creating generalized physical models for underwater ocean acoustics is a difficult task due to the complex multi-physics phenomena involved, the wide range of physical scales, and the uncertainties associated with numerous environmental parameters \cite{james2005probability, james2011pekeris}.
This research focuses on building a data-driven physics-based machine learning algorithm designed to forecast noise originating from distributed sources at the expected locations of marine mammals in nearly real-time.

Machine learning has experienced a resurgence in recent years, largely due to the impressive accomplishments of deep learning (DL) models in diverse fields. \cite{lecun2015deep}. 
One area experiencing a notable resurgence is reduced-order modeling, which utilizes deep learning models to approximate complex physical systems \cite{deo2022predicting, deo2023combined, gao2024finite}. 
Among the numerous deep learning models available, deep neural networks stand out as a highly effective technique for representing and modeling physical phenomena, as exemplified by their successful application in modeling complex systems like the Navier-Stokes equations~\cite{yang2016data}.
We propose utilizing a data-driven method based on deep learning to construct a model of underwater sound propagation. The data-driven approach relies solely on the data, independent of the method used to generate the data. As a result, this can effectively approximate the underlying ocean acoustic environment across diverse environmental conditions and varying ocean bathymetry, utilizing data acquired from experiments or computational solvers. Moreover, these data-driven models can distill a much lower-dimensional representation of the system from high-dimensional physical data, enabling real-time predictions.

Recently, a variety of data-driven models, including DL models, have been utilized in underwater acoustics for classifying and localizing ship-generated sound sources based on their acoustic characteristics in both shallow and deep ocean environments \cite{niu2017ship, huang2018source, chi2019sound, ferguson2021multitask}. Additionally, DL models have been increasingly employed to investigate shallow water wave propagation \cite{fotiadis2020comparing,deo2022predicting}.  However, some of these approaches explicitly rely on partial differential equations \cite{borrel2021physics}, making them less suitable for handling high-dimensional physical data and complex systems where the exact governing equations and underlying physical processes may be unknown.
Of particular interest, Mallik et al. applied a convolutional recurrent autoencoder network (CRAN) for predicting transmission loss \cite{mallik2022predicting}.
CRAN \cite{bukka2021assessment,deo2022predicting} operates similarly to numerical discretization methods. It encodes high-dimensional physical data into a latent space and subsequently propagates it along the temporal dimension, analogous to discretization methods that discretize physical space and use time integration for solution marching.
 This methodology addresses the challenges related to knowing the underlying differential equation and mitigates the computational burden associated with numerical methods.
However, CRAN-type architecture struggles to predict far-field noise due to their autoregressive prediction where information from the previous step is used to predict the next step. Since these architectures lack information on the whole ocean bottom's geometry, they are not able to predict underwater radiated noise at distant locations with accuracy.

To tackle these challenges, we present a novel range-dependent conditional convolutional neural network (RC-CAN) designed to predict transmission loss accurately based on input ocean bottom geometry. Our method aims to achieve a single-shot prediction, thus alleviating the limitations of autoregressive prediction methods.
The RC-CAN model is a composite encoder-decoder framework, which transforms the input ocean bathymetry mesh using an encoder network and uses a decoder network to decode the transmission loss on the input mesh. The RC-CAN is employed for learning the spatial distribution of far-field transmission loss from a point source in a two-dimensional underwater ocean environment with varying bathymetry. We consider a wide range of varying ocean bathymetry to demonstrate the generalized learning capacity of the RC-CAN model. 

For the first time, we approach underwater radiated noise modeling as a continual learning task, leveraging prior work in the field \cite{10444954, rebuffi2017icarl}. This approach allows us to adjust to changing ocean bathymetry while retaining previously learned information. 
We introduce a novel replay-based training method for the RC-CAN model \cite{rolnick2019experience}, marking the first application of a continual learning framework in the context of dynamical systems prediction.
The replay-based training method entails storing a subset of data from past tasks or conditions and retraining the model on this subset \cite{tiwari2022gcr}. This strategy ensures that the model maintains memory of earlier tasks, mitigating catastrophic forgetting, a common issue in sequential learning scenarios where the model forgets previous information when trained on new data \cite{kirkpatrick2017overcoming, bhatt2024preventing}.
By employing a replay-based strategy, the RC-CAN model can sequentially train on new ocean bathymetry data while retaining knowledge from prior training instances \cite{ke2021achieving}. This successful implementation opens up avenues for future applications of data-driven models in real-time forecasting of far-field underwater noise transmission across diverse parameters and bathymetry. Integration of this method into IGMV noise mitigation systems holds promise for continual operation adaptation and reduction of the impact on marine mammals.

The rest of the paper is structured as follows. Section 2 covers mathematical preliminaries and elaborates on our proposed RC-CAN network. Section 3 describes the  replay training strategy used for our network architecture. In Section 4, we perform a numerical analysis focusing on predicting far-field underwater radiated noise using RC-CAN over a wide variety of varying bathymetry. The paper concludes in Section 5, where we provide a brief discussion of our findings and propose potential avenues for future research.

\section{\label{sec:2} Mathematical Formulation}

This section begins with a concise overview of data-driven reduced-order modeling in the context of underwater radiated noise. 
Following this, we introduce our range-dependent conditional convolutional neural network.
\subsection{Data-driven reduced-order modeling}
Underwater noise transmission is governed by the propagation of acoustic pressure generated from sound sources. 
The propagation of acoustic pressure is modeled using the wave equation for pressure perturbation given by:
\begin{equation}
    \label{eq:pressure_waveEQN}
\rho \nabla \cdot\left(\frac{1}{\rho} \nabla p\right)-\frac{1}{c^2} \frac{\partial^2 p}{\partial t^2}=0,
\end{equation}
where $p$ and $\rho$ are pressure and density perturbations. If the density is constant in space, Eq. \ref{eq:pressure_waveEQN} can be replaced by the standard form of the wave equation,
\begin{equation}
    \label{eq:std_pressure_waveEQN}
\nabla^2 p-\frac{1}{c^2} \frac{\partial^2 p}{\partial t^2}=0 .
\end{equation}
Since the coefficients of the two differential operators in Eq. \ref{eq:std_pressure_waveEQN} are independent of time, the dimension of the wave equation can be reduced using frequency–time Fourier transform leading to Helmholtz equation in Cartesian coordinates $\mathbf{x}=(x, y, z)$, given by:
\begin{equation}
    \label{eq:helmohotz_eqn}
    \nabla^2 p+\frac{\omega^2}{c^2(\mathbf{x})} p=-\delta\left(\mathbf{x}-\mathbf{x}_0\right),
\end{equation}
where $c(\mathbf{x})$ is the sound speed and $\omega$ is the angular frequency of the source located at $\mathbf{x}_0$. To establish the computational domain, it is necessary to specify the boundary conditions, which are expressed as:
\begin{equation}
    \label{eq:helmohotz_eqn_BC}
     p(\partial\mathbf{x}; \omega) = g(\partial\mathbf{x}; \omega),
\end{equation}
where $\partial\mathbf{x}$ is the boundary of the domain and Eq. \ref{eq:helmohotz_eqn_BC} represents the boundary conditions of the PDE.

\begin{figure}[H]
    \centering
    \includegraphics[width=0.99\linewidth]{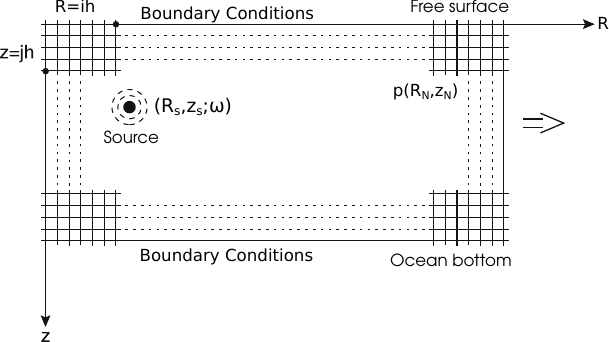}
    \caption{A schematic of finite-difference approximation on a rectangular mesh.}
\label{fig:FDM}
\end{figure}

In a simplified two-dimensional representation of the ocean environment, the acoustic pressure can be expressed in Cartesian coordinates as $p(R,z;\omega)$. Here, $p(R,z;\omega)$ depends on the range $R$, depth $z$, and the frequency of the emitted signal $\omega$. 
Many numerical techniques, such as the finite-difference method (FDM), discretize the space $(R,z)$ to obtain the solution. FDM solves the Helmholtz equation on the discretized grid $(R_N,z_N)$ to compute the pressure $p(R_N, z_N)$ at the discretized locations. 
These methods incorporate ocean bathymetry information into the boundary conditions of the partial differential equations (PDEs). Figure \ref{fig:FDM} represents a rectangular finite-difference mesh for such applications.

The acoustic energy or intensity $\left(I(z,R)\text{, and  I} \propto p^2\right)$ reduces significantly with the distance from the source and this is referred to as transmission loss (TL). This loss can be measured in decibels by comparing the acoustic pressure $p(R,z)$ at a receiver location $(R,z)$ with the reference signal $p_0$ emitted by the source.

\begin{equation}
    \label{eq:TL_comp}
    \mathrm{TL} (R,z) = -10 \log \frac{I(R,z)}{I_0} = - 20 \log \frac{p(R,z)}{p_0}.
\end{equation}
Hence the transmission loss is a measure of the drop in far-field sound pressure level from various sources of URN. For this study, we want to build a reduced-order model that predicts the far-field transmission loss on the input mesh. We can employ the deep neural network to build such a reduced-order model given by:
\begin{equation}
    \label{eq:DL_ROM}
    \mathrm{TL} (R_N,z_N) = \mathcal{NN} (R_N, z_N ; \theta), 
\end{equation}
here, $(R_N,z_N)$ represents the input mesh containing data regarding receiver locations and ocean bathymetry. $\theta$ denotes the trainable parameters of the deep neural network. Therefore, the objective of this study is to learn the operator $\mathcal{NN(.)}$ that transforms the input mesh $(R_N,Z_N)$ with bathymetry information to transmission loss $\mathrm{TL} (R_N,z_N)$ using a data-driven approach. The subsequent section introduces the range-dependent conditional convolutional neural network, which is utilized to construct this reduced-order model.

\subsection{Range-dependent conditional convolutional neural network}

\begin{figure*}[ht]
    \centering
    \includegraphics[width=\textwidth]{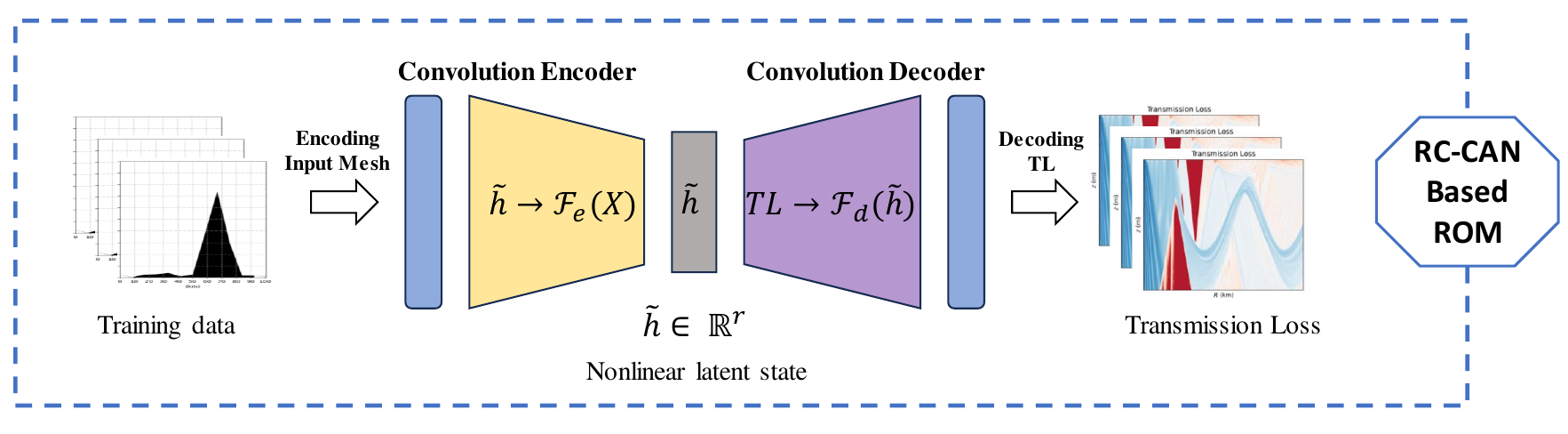}
    \caption{An illustration of the proposed range conditional convolutional network. }
\label{fig:RC-CAN_ROM}
\end{figure*}

To develop a reduced-order model for predicting acoustic transmission loss, we will utilize a range-dependent conditional convolutional neural network based on a 2D convolutional neural network architecture. 
First, a stack of convolutional and fully connected neural networks transforms the input ocean bathymetry mesh into a latent space.
Subsequently, a 2D convolution decoder network will be employed to learn a mapping from the low-dimensional manifold to a high-dimensional space. This approach aims to capture the relationship between ocean bathymetry and far-field transmission loss by applying convolutional kernels to the input mesh representing ocean bathymetry, with the decoder network learning the transmission loss on the input mesh.
Figure \ref{fig:RC-CAN_ROM} depicts the process of encoding input mesh into latent space and then decoding the transmission loss on the input mesh.
The encoder and decoder can be expressed as:

\begin{subequations}
\begin{align}
\tilde{h}_r=\boldsymbol{\Psi}_{E}\left(\mathbf{R_N,z_N};\theta_{E}\right)&,\label{eq:enc}\\
\hat{\mathrm{TL}}(R_N,z_N)=\boldsymbol{\Psi}_{D}\left(\tilde{h}_r;\theta_{D}\right)&,\label{eq:dec}
\end{align}
\end{subequations}
where $\hat{\mathrm{TL}}(R_N,z_N)$ denotes the approximation of the transmission loss on the input mesh $ (R_N,z_N)$. The ocean bathymetry and domain are masked on the input mesh $ (R_N,z_N)$. $\boldsymbol{\Psi_{E}(.;\theta_{E})}$ stands for the encoder network that maps the input ocean bathymetry to a low-dimensional manifold $\mathbf{\tilde{h}_r}$, and $\theta_{E}$ is the parameter of the encoder network. $\boldsymbol{\Psi_{D}(.;\theta_{D})}$ denotes the decoder network that maps the low-dimensional data back to the high-dimensional physical state, and $\theta_{D}$ is the parameter of the decoder. 
By backpropagating the $L_2$ norm of error, the weights of the neural network $(\theta_E,\theta_D)$, can be trained.

We outline key characteristics of convolutional neural networks that make them suitable for reduced-order modeling in transmission loss prediction. Firstly, underwater radiated noise involves the spread of a disturbance from its source, with a pattern that propagates through the domain. The translational invariance property of convolutional neural networks is beneficial for modeling this propagation \cite{bronstein2017geometric}. Additionally, convolutions are well-suited for learning ocean bathymetry features as they can effectively capture local features using different convolutional kernels.

The 2D convolutional network employed in this study consists of two main components: an encoder and a decoder path. The encoder path follows a standard convolutional network architecture, comprising four padded convolutions, each followed by batch normalization and Leaky ReLU activation. Subsequently, a max pooling operation is applied to down-sample the data. Mathematically, the encoder path is represented by Eq. \ref{eq:enc}.
The decoder path, represented by Eq. \ref{eq:dec}, involves up-sampling the feature map followed by four transpose convolutions (up-convolution), each also followed by batch normalization and Leaky ReLU activation. The final layer uses a 2D convolution kernel with one feature channel to align with the input data. Overall, the network comprises 8 convolutional layers. 

By utilizing the encoder and decoder network, we can construct the reduced-order model or learn the $\mathcal{NN(.)}$ operator as a composition of the encoder and decoder, expressed as:
\begin{equation}
    \label{eq:NN_operator}
    \mathcal{NN(.)} =\boldsymbol{\Psi}_{D} \circ \boldsymbol{\Psi}_{E}(.)  . 
\end{equation}


The training of this convolutional network involves finding the parameters that minimize the expected reconstruction error over all training examples given by
\begin{equation}
{\theta}_{E}^{*},{\theta}_{D}^{*}=\arg \min _{{\theta}_{{E}},{\theta}_{D}}\mathcal{L}[\mathrm{TL}(R_N,z_N),\Psi_{D}(\Psi_{E}({R_N,z_N};\theta_{E});\theta_{D})],
\label{eqn: autoencoder loss}
\end{equation}
where $\mathcal{L}[\mathrm{TL}(R_N,z_N),\Psi_{D}(\Psi_{E}({R_N,z_N};\theta_{E});\theta_{D})]$ is a loss function in the $L^{2}$ norm,  which minimizes the difference between the reconstruction $\left(\Psi_{D}(\Psi_{E}({R_N,z_N};\theta_{E});\theta_{D})\right)$ and the ground truth $\left(\mathrm{TL}(R_N,z_N)\right)$. The training of RC-CAN is discussed in detail in the subsequent section.
\section{\label{sec:3}Replay-based Training Strategy of RC-CAN}
In this section, we turn our attention to the details of creating the training dataset and the associated training procedure. We curate a dataset containing seven different ocean bathymetry profiles, which can be categorized into three groups. The first group includes an idealized representation of seamounts using triangles. 
Seamounts are underwater mountains that rise from the ocean floor but do not reach the water’s surface.
The second group comprises wedge profiles, while the third group consists of bathymetry profiles sampled from Dickin's seamount location in the Northeast Pacific Ocean.

\begin{figure}[h]
    \centering
    \includegraphics[width=0.99\linewidth]{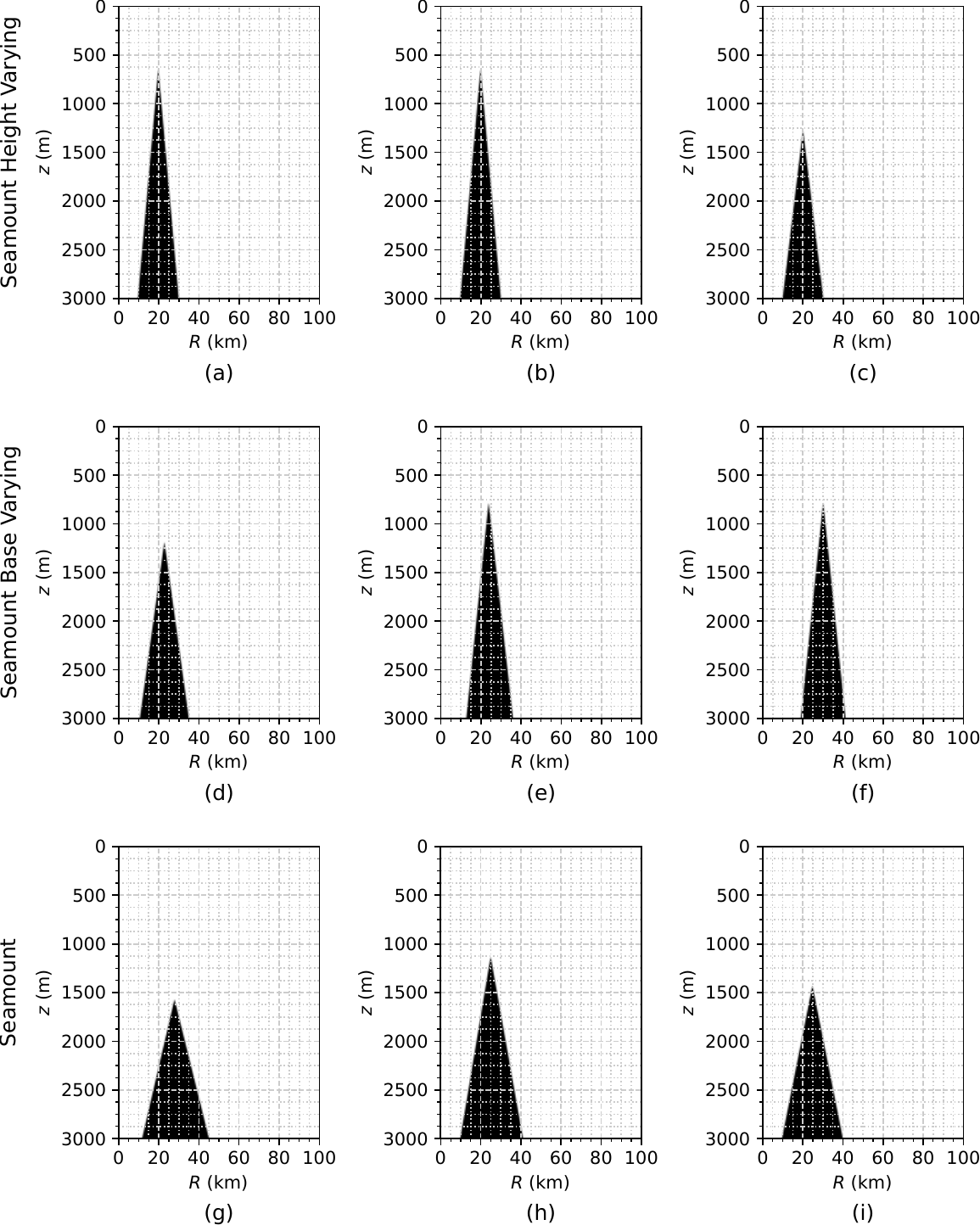}
    \caption{Idealized seamount profiles. (a-c): fixed base width (20 km) and peak depth (500 m to 1600 m). (d-f): peak depth (800 m or 1200 m) and base width (5 km to 40 km). (g-i): randomly sampled peak depths and base widths.}
\label{fig:SeaMount}
\end{figure}

The idealized seamount profiles assume a flat bottom bathymetry at a depth of 3000 m, except for the area where the seamount rises. These profiles are divided into three sub-categories. The first subgroup assumes a fixed base of the seamount, which is 20 km wide and located 10 km from the source. The peak of the seamount is located at a range of 20 km from the source, with the depth of the peak varying from 500 m to 1600 m. To create the dataset for this subgroup, 75 values of peak depth are randomly sampled between these heights.
The second subgroup assumes that the depth of the peak is either 800 m or 1200 m, while the base width of the seamount varies randomly from 5 km to 40 km. For this subgroup, 35 values are randomly sampled between these points.
The third subgroup represents a generalized case where the peak depth and width are randomly sampled. These three subgroups are illustrated in Figure \ref{fig:SeaMount}.

\begin{figure}[h]
    \centering
    \includegraphics[width=0.99\linewidth]{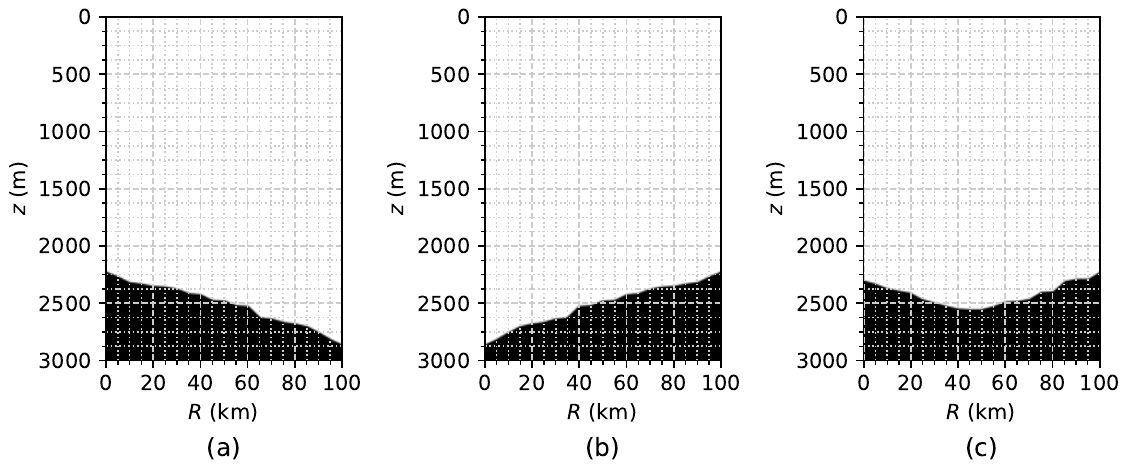}
    \caption{Wedge bathymetry profiles. (a) Depth decreasing profile, (b) increasing depth profile, and (c) decreasing then increasing depth profile. }
\label{fig:WedgeInput}
\end{figure}

\begin{figure}[h]
    \centering
    \includegraphics[width=0.99\linewidth]{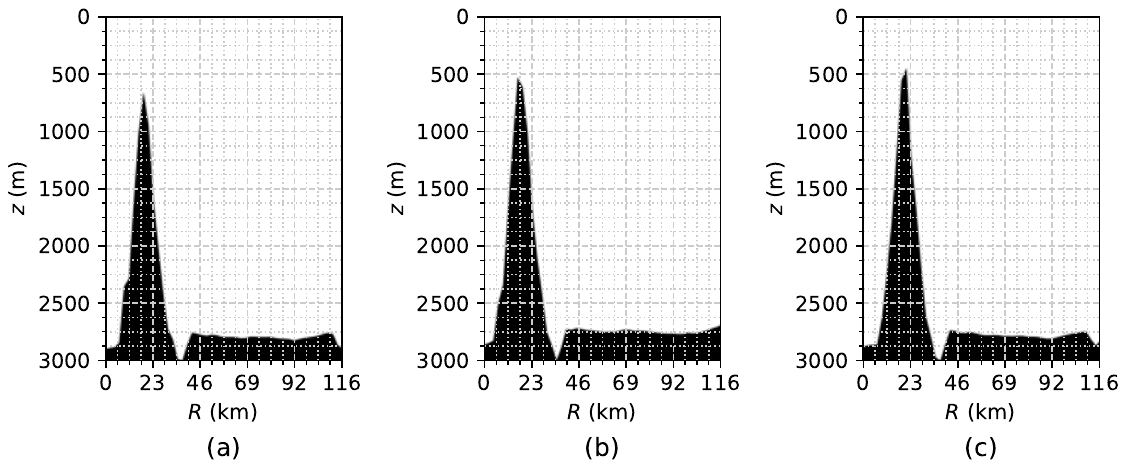}
    \caption{An image depicting a cross-sectional view of the Dickins seamount in the range-depth plane. }
\label{fig:DickinsInput}
\end{figure}

The second group comprises wedge profiles, characterized by gradual changes in ocean bathymetry depth, either increasing or decreasing.  A special case within the wedge profile includes both decreasing and increasing profiles within the data range. Figure \ref{fig:WedgeInput} shows the wedge bathymetry profiles. The third group consists of bathymetry profiles sampled from the location of Dickin's seamount in the Northeast Pacific. 
Figure \ref{fig:DickinsInput} depicts the bathymetry of Dickin's seamount. 
If the network architecture inherently possesses a strong implicit bias to capture the physics of underwater radiated noise, then having a small sample size of bathymetry instances from this location could be sufficient for training RC-CAN in a realistic ocean environment. 
Therefore, we sampled only 10 bathymetries from the Dickins seamount location for training.

Each bathymetry profile corresponds to 2049 receiver locations along the depth and 1408 receiver locations along the range. The depth of the ocean domain is fixed at 3000 m, enabling the examination of deep ocean sound propagation. Furthermore, the range considered for this study spans 100 km in the domain.
The input for training consists of a mesh that covers the receiver locations, with the mesh dimensions specified as $1408 \times 2049$. In this mesh representation, ocean bathymetry is denoted by a value of 1, while the ocean itself is represented by a value of 0. This mesh configuration is visually depicted in Figure \ref{fig:SeaMount}.

The training dataset $\mathcal{D = \{D_\text{1},...,D_\text{7}} \}$ comprises three groups: $\{\mathcal{D_\text{1}, D_\text{2},  D_\text{3}}\}$ blong to idealized seamount group, $\{\mathcal{D_\text{4}, D_\text{5},  D_\text{6}}\}$ represent wedge profiles, and $\mathcal{D_\text{7}}$ depicts the Dickins seamount data. We denote the task of predicting transmission loss for the dataset  $\mathcal{D_\text{i}}$ as Task $\mathcal{\text{i}}$. 


When the model is trained on the entire dataset $\mathcal{D}$, it does not perform well in generalization or in predicting transmission loss for the individual tasks. To address this, the model can be trained sequentially on Task ${1}$, to Task ${7}$. Initially, training the model solely on Task $1, 2, \text{and } 3$ yields exceptional performance in predicting transmission loss for the idealized seamount. However, training on Task ${4}$ causes the weights to be optimized for Task ${4}$, resulting in suboptimal performance on Task $1, 2, \text{and } 3$. This phenomenon is known as catastrophic forgetting. 

A continual learning system must exhibit both plasticity \cite{ostapenko2019learning}, involving the acquisition of new knowledge, and stability, involving the retention of old knowledge \cite{kong2022balancing}. Catastrophic forgetting arises when stability breaks down, leading to new experiences overriding previous ones. In the brain, it is widely believed that replaying past experiences helps alleviate forgetting \cite{rolnick2019experience}.
To address this issue, a replay buffer dataset $\mathcal{K}$ can be created. After completing training on Task $i$, the model is re-trained for a fixed number of epochs on $\mathcal{K_\text{i}}$. The replay buffer $\mathcal{K_\text{i}}$ contains an exemplar set $\{\mathcal{P_\text{1}},...,\mathcal{P_\text{i-1}}\}$, where $\mathcal{P_\text{i}}$ consists of representative examples from the dataset $\mathcal{D_\text{i}}$. Several methods exist for selecting the exemplar set, such as the gradient coreset replay buffer, which is a strategy for selecting and updating the replay buffer using a carefully designed optimization criterion \cite{tiwari2022gcr}. In this study, we randomly sample examples \cite{brahma2018subset} from the dataset $\mathcal{D_\text{i}}$ to create the exemplar set $\mathcal{P_\text{i}}$. The complete training algorithm is presented in the subsequent sections.

For improved neural network training and to prevent the over-saturation of any particular feature, the training transmission loss for each profile are scaled appropriately as follows: 
\begin{align}
\overline{\mathrm{TL}}_{\text{Train}}&=\frac{{\mathrm{TL}}_{\text{Train}} - {\mathrm{TL}}_{\text{Train}_{,\mu }}}{{Std.Dev.(\mathrm{TL}}_{\text{Train}})},
\end{align}
where  $\overline{\mathrm{TL}}_{\text{Train}}$ represents the transmission loss data that has been normalized to have a mean of zero and a standard deviation of one. This normalized training set is then utilized for training the model.

The range conditional convolution neural network makes predictions in a single shot on a given input bathymetry mesh. 
The input dimensions are $1408 \times 2049$.
We want to determine the model parameters  $\theta^* = \left\{\theta_E^*,\theta_D^*\right\}$ such that they minimize the following expected error between the model prediction and the ground truth data 
\begin{equation}
\begin{aligned}
\mathcal{L}\left(\tilde{\mathrm{TL}}, \mathrm{TL}\right)
=\frac{1}{N_m}\sum_{j=1}^{N_m}\left(\left\|\tilde{\mathrm{TL}}_{j}-{\mathrm{TL}}_{j}\right\|_{2}^{2} \right),
\end{aligned}
\label{eqn:loss}
\end{equation}
where $\tilde{\mathrm{TL}}$ and $\mathrm{TL}$ are the model prediction and the ground truth respectively. While $N_m$ denotes the various ocean bathymetry profiles.

In the present work, we employ a cosine annealing warm restart learning rate scheduler. This scheduler utilizes the cosine function as the learning rate annealing function and periodically resets or increases the learning rate \cite{loshchilov2016sgdr}.
We use the AdamW optimizer \cite{loshchilov2018decoupled}, a version of the adaptive gradient algorithm Adam. Algorithm \ref{alg:alg1} provides a complete training procedure for our architecture.


\begin{algorithm}[h]
\RestyleAlgo{ruled}
\SetKwInput{KwOutput}{Output}

\caption{Range-Dependent Conditional Convolutional Network Training}

\SetAlgoLined

\KwIn{Dataset: $\mathcal{D}$, Parameters: $\theta$, Patience: $N_p$, Epochs: $N_{\text{epochs}}$, Batch size: $N_b$, Replay Epochs: $ N_\text{replay}$ }

\KwOutput{$\{\theta^{*}  := (\theta_{E}^{*},\theta_{D}^{*}\}$)}
 
 Initialize Parameters: $\theta$\, and Replay Buffer: $\mathcal{K}=\emptyset$\;

 $m \leftarrow 0 / / \text { The counter for early stopping }$\;
 
 \For{ $\mathcal{D}_\text{i} \in \mathcal{D}$}{ 
 
 \While{ epoch $<$ $N_{\text{epochs}}$ and $m < N_p$ }
 {
 $\left((R_N,z_N)_{\text{Train}},\mathrm{TL}_{\text{Train}}\right) \in \mathcal{D_\text{i}}$\;

 Randomly sample batch of input bathymetry from training data: ${(R_N,z_N)}_{\text{Train}}^{b} \subset {(R_N,z_N)}_{\text{Train}}$\;
 
 Forward pass: ${{\tilde{\mathrm{TL}}}}^{b} \leftarrow \Psi_{D}\left(\Psi_{E}\left({(R_N,z_N)}_{\text{Train}}^{b};\theta_{E}\right);\theta_{D}\right)$, ${\tilde{\mathrm{TL}}}^{b} \in \mathbb{R}^{N_b \times R_N \times z_N}$\;

Calculate loss $\mathcal{L}$ via Eq.~(\ref{eqn:loss})\;

Estimate gradients $\hat{\mathbf{g}}$ using automatic differentiation\;

Update parameters: $\theta \leftarrow A D A M W(\hat{\mathbf{g}})$\;

\eIf{$\mathcal{L}$ $\leq$ ${l_b}$}
{
$\text{Save the model parameters}$\;
$m \leftarrow 0$\;
 /* lb is best loss */\\
${l_b}$ $\leftarrow \mathcal{L}$\;
}
{
$m \leftarrow m+1$
}
}

$\mathcal{P}_{i-1} \leftarrow \text {ConstructExemplarSet}\left(\mathcal{D}_{i-1}\right)$\;
$\mathcal{K}_i \leftarrow \text {ConstructReplayBuffer}\left(\mathcal{P}_1, ..., \mathcal{P}_{i-1}\right)$\;

\While{ epoch $<$ $N_{\text{replay}}$}
{
$\left((R_N,z_N)_{\mathcal{K}_i},\mathrm{TL}_{\mathcal{K}_i}\right) \in \mathcal{K}_i$\;
${(R_N,z_N)}_{\mathcal{K}_i}^{b} \subset {(R_N,z_N)}_{\mathcal{K}_i}$\;
Forward pass: ${{\tilde{\mathrm{TL}}}}^b \leftarrow \Psi_{D}\left(\Psi_{E}\left({(R_N,z_N)}_{\mathcal{K}_i}^{b};\theta_{E}\right);\theta_{D}\right)$, ${\tilde{\mathrm{TL}}}^b \in \mathbb{R}^{N_b \times R_N \times z_N}$\;
Calculate loss $\mathcal{L}$ via Eq.~(\ref{eqn:loss})\;
Estimate gradients $\hat{\mathbf{g}}$ using automatic differentiation\;
Update parameters: $\theta \leftarrow A D A M W(\hat{\mathbf{g}})$\;
}
}

Updated parameters: $\{\theta^{*}  := (\theta_{E}^{*},\theta_{D}^{*})\}$

\label{alg:alg1}

\end{algorithm}

Once the model is trained, predicting transmission loss becomes straightforward. The network is used to forecast transmission loss on input bathymetry meshes using testing data sets. These data sets are created by segregating some data from the training set. 
Given that the true and predicted solutions in this context can be represented in a rasterized format on a Cartesian grid, the accuracy of the predicted solutions was assessed using the Structural Similarity Index Measure (SSIM). 
SSIM is a statistical measure employed to compare two datasets of the same dimensions with rasterized representations. 
Further details on SSIM can be found in Ref. \cite{wang2004image}. 
An SSIM value of 1.0 between two inputs indicates perfect similarity, while a value of 0 implies no similarity.
 Next, we turn our attention to a detailed assessment of the model.


\section{Test scenario for RC-CAN based learning}
\subsection{Test problem}

\begin{figure}[ht]
\centering
\includegraphics[width=0.99\linewidth]{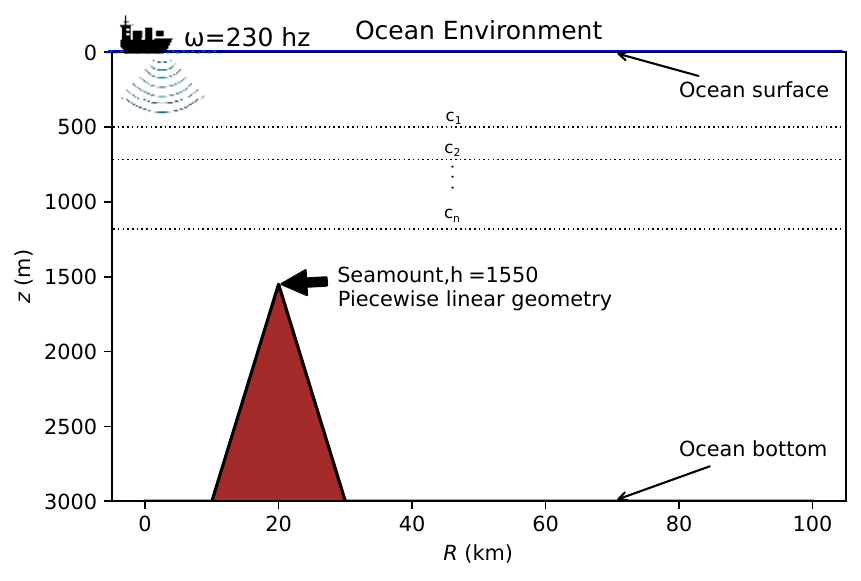}
\caption{An illustration of ocean domain with range-dependent bathymetry}
\label{fig:range-dependent_ocean_domain}
\end{figure}

To illustrate the learning capability of RC-CAN in predicting ocean acoustic transmission loss, we examine an oceanic environment with depth-dependent sound speed following Munk's sound speed profile \cite{jensen2011computational}. The computational domain for the ocean is depicted in Fig. \ref{fig:range-dependent_ocean_domain}. In this scenario, the domain spans 100 kilometers in range and has a depth of 3000 meters, featuring a piece-wise linear bathymetry profile. 
This research aims to employ a data-driven method to learn transmission loss from 230 Hz point sources at a depth of 18 meters, with ocean bathymetry varying along the range. 
We specifically consider transmission losses primarily attributed to geometric spreading and reflections from the top and bottom surfaces, excluding losses due to volume attenuation. The ocean bottom is treated as a rigid boundary, while the ocean surface is regarded as a fully reflective pressure release boundary \cite{jensen2011computational}. Consequently, while reflections occur from the ocean surface and bottom, scattering losses from the ocean bottom are deemed negligible.
%
\subsection{Generation of ground truth for far-field transmission loss}

In the physical conditions outlined in this study, the transmission of underwater radiated noise can be viewed as a linear wave propagation process. Consequently, it can be mathematically represented in the frequency domain using the Helmholtz equation in Cartesian coordinates $\bm{\mathrm{x}}$.
\begin{equation}
    \label{eq:Helmholtz_eq}
 \nabla^2 p+\frac{\omega^2}{c^2(\mathbf{x})} p=-\delta\left(\mathbf{x}-\mathbf{x}_0\right), \quad \bm{\mathrm{x}} \in \Omega,
\end{equation}
subject to the boundary conditions. In the context of a two-dimensional ocean domain $\Omega$, the variable $p$ denotes the far-field acoustic pressure perturbation originating from a source located at $\bm{\mathrm{x_0}}$. We can express $\bm{\mathrm{x}}$ as $\left( R, z \right)$, and $\nabla^2$ as $\frac{\partial^2}{\partial R^2} + \frac{\partial^2}{\partial z^2}$, which is a valid assumption without loss of generality. Here, $c\left(\bm{{x}}\right)$ represents the speed of sound, $R$ denotes the direction along the range of the ocean domain from a reference point, and $z$ indicates the depth from the ocean surface. Equation \ref{eq:Helmholtz_eq} can be approximately solved using ray tracing techniques, enabling the analysis of a broad range of depth and range-dependent variations in ocean environment parameters (e.g., sound speed, bathymetry, etc.).
The ray method is especially beneficial for tackling range-dependent problems at high frequencies, situations where normal mode or parabolic models are not practical options.

While ray tracing techniques ignore the lower frequency solutions of the Helmholtz equation \cite{jensen2011computational}, established computational methods for ray tracing \cite{porter1987gaussian,porter1994finite} have demonstrated the ability to yield reasonably accurate results even at frequencies as low as 50 Hz \cite{bellhop}. Ray tracing approaches can cause anomalies such as sharp caustic and shadow zones in ray amplitude calculations. However, the use of geometric beams in ray-beam tracing solvers can mitigate the occurrence of caustics to acceptable levels, as documented in Ref. \cite{porter1994finite}.

In this study, we calculate the far-field acoustic pressure levels considering geometric spreading, ocean surface, and bottom reflections using BELLHOP Gaussian beam tracing solver \cite{bellhop}. BELLHOP is employed to compute the acoustic pressure at evenly spaced receiver locations along the depth ($z$) and range ($R$) throughout the prescribed ocean domain, for a given point source depth ($z_s$). These acoustic pressure amplitudes are utilized to derive transmission loss, and a snapshot of the transmission loss distribution is extracted on a uniformly discretized $R-z$ plane for the ocean domain depicted in Fig. \ref{fig:range-dependent_ocean_domain}.  
Our objective is to develop a generalized learning model for far-field transmission loss in bathymetry-varying scenarios using the RC-CAN model. This involves training the model on various bathymetry profiles and demonstrating transmission loss predictions for any bathymetry not included in the training set.

It's worth noting that although BELLHOP computes acoustic pressure, we train the RC-CAN on transmission loss data obtained by post-processing the intensity of the pressure signal (according to Eq. \ref{eq:TL_comp}). This choice is made because acoustic pressure varies over several orders of magnitude in such a large domain, and deep learning networks cannot be effectively trained on such data. Conversely, the transmission loss of acoustic pressure is easier to interpret and more suitable for training the network.

\section{\label{sec:6} Results}

In this section, we demonstrate the capability of the proposed RC-CAN architecture to make data-driven predictions of transmission loss, particularly in far-field scenarios. We also showcase its ability to generalize across various ocean bathymetry conditions. To validate the reliability and accuracy of our methodology, we will test it in scenarios involving the prediction of underwater radiated noise over idealized seamounts, wedge profiles, and Dickins seamount. Various components of the RC-CAN architecture were trained with PyTorch \cite{NEURIPS2019_9015} library.

\subsection{Training of RC-CAN}

\begin{figure}[ht!]
\centering
\subfigure[Seamount Height Varying ($\mathcal{D}_{\text{1}}$)]{\includegraphics[width=0.99\linewidth]{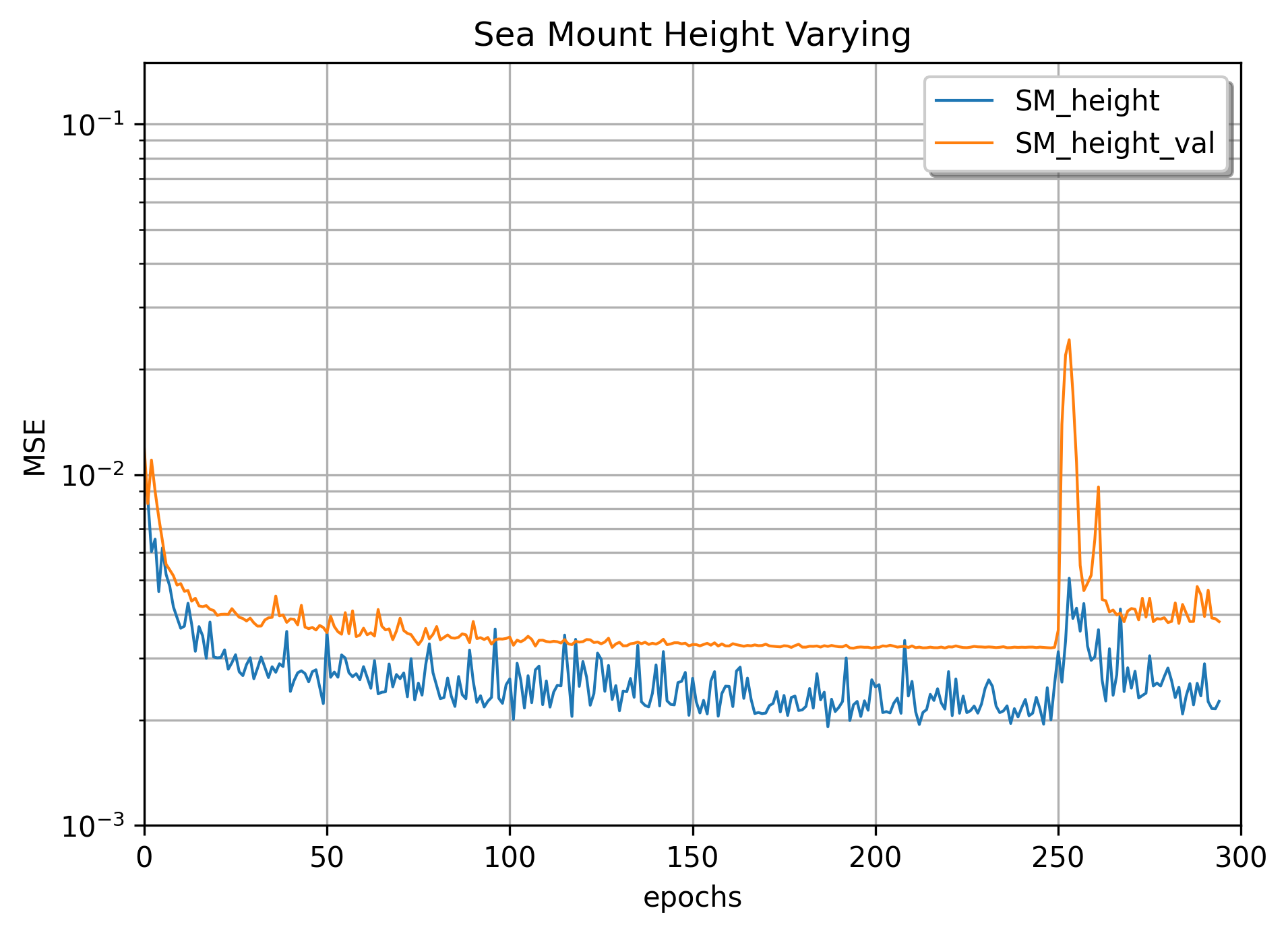}}
\subfigure[Seamount Base Varying ($\mathcal{D}_{\text{2}}$)]{\includegraphics[width=0.99\linewidth]{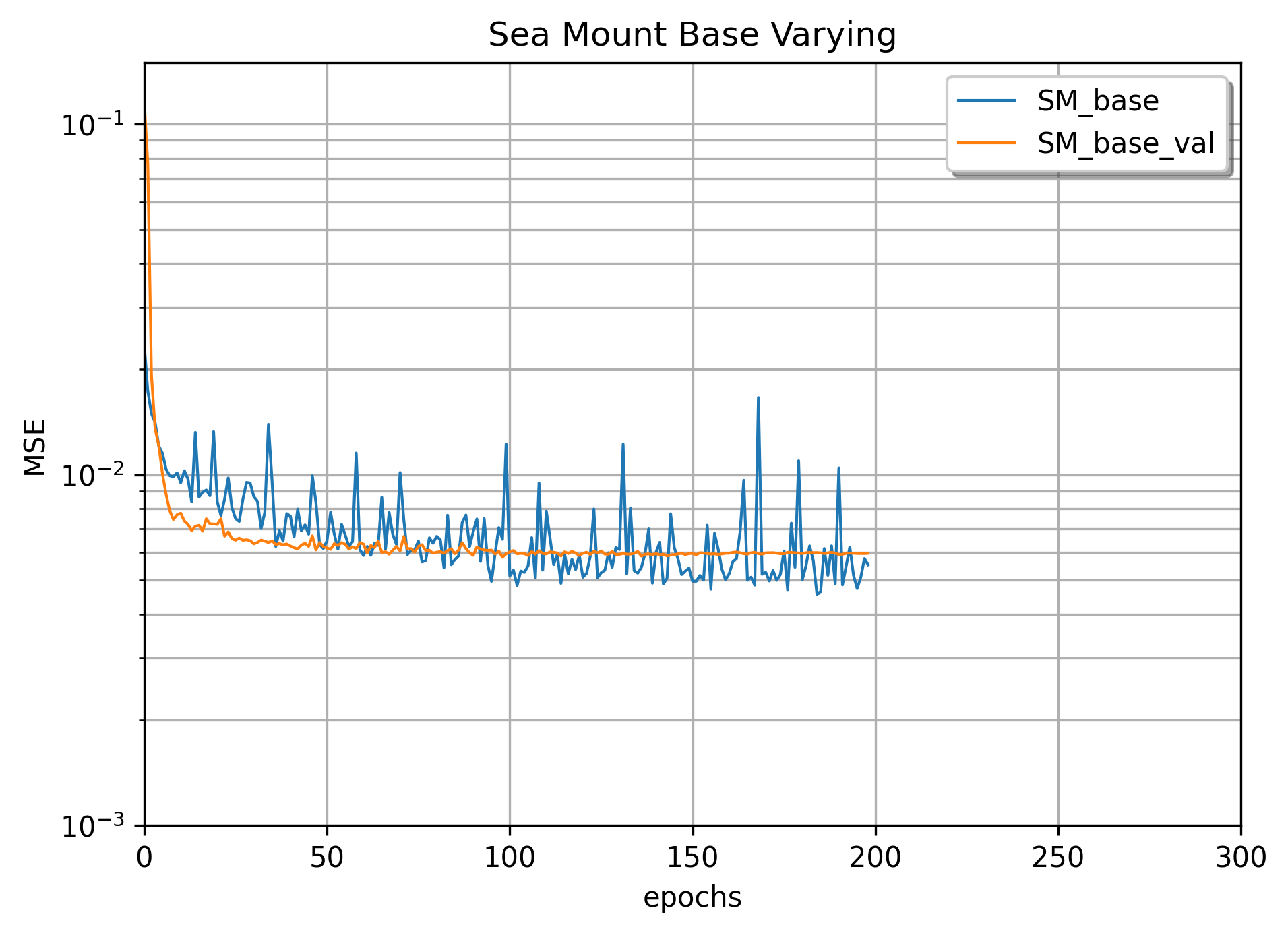}}



\caption{Training and validation loss plot for dataset $\mathcal{D} = {\mathcal{D}_{\text{1}}, \& \mathcal{D}_{\text{2}}}$}
\label{fig:lossvsepoch}
\end{figure}

The RC-CAN was trained using BELLHOP-generated data to account for transmission loss distribution across a range of varying bathymetry conditions. The training process involved sequential training on different datasets, starting with an ideal seamount dataset, followed by wedge profiles, and concluding with the Dickins seamount dataset. Fig. \ref{fig:lossvsepoch} illustrates the progression of mean-squared error during training and validation across these datasets. Notably, the model required the greatest number of epochs to train on the dataset with varying sea mount heights. Training patience (i.e., the number of epochs to wait before early stop if no progress on validation set) of 100 epochs was set, indicating that if the validation mean-squared error did not decrease within these epochs, the training process would stop. Subsequently, the model underwent re-training on a replay buffer dataset $\mathcal{K}{\text{i}}$ to preserve knowledge from previous tasks. Transfer learning was applied when transitioning to a new dataset $\mathcal{D}{\text{i}}$, significantly reducing the required number of epochs for convergence. For instance, training on a dataset with varying sea mount bases only necessitated 200 epochs, compared to approximately 300 epochs needed for the dataset with varying sea mount heights (as depicted in Fig. \ref{fig:lossvsepoch}).

\subsection{Underwater Radiated Noise}

\begin{figure}[ht!]
    \centering
    \includegraphics[width=\linewidth,keepaspectratio]{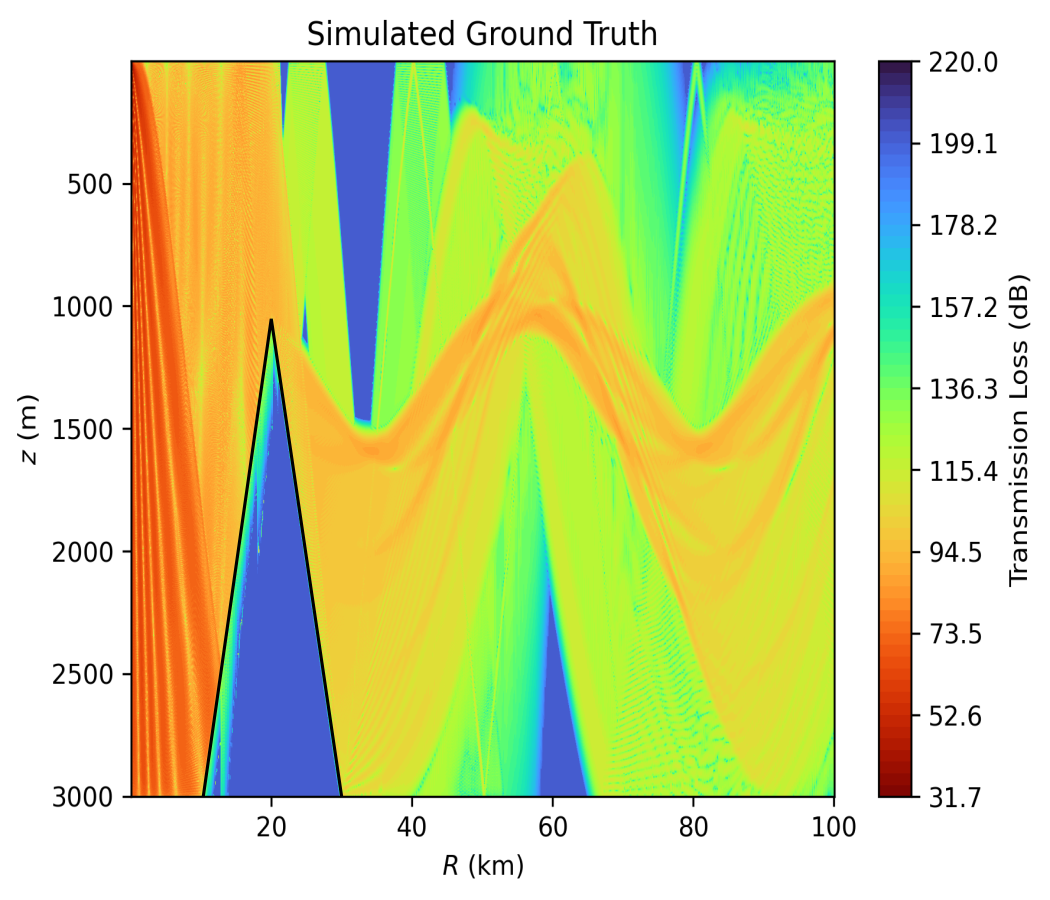}
    \caption{Transmission loss distribution obtained via BELLHOP for a source depth of 18 meters over ideal seamount}
    \label{fig:tl_soln_zs18}
\end{figure}

The transmission loss distribution computed over the domain via BELLHOP for a source at a depth $z_s=18$ meters is shown in Fig. \ref{fig:tl_soln_zs18}. 
BELLHOP and most other routine rays/beam tracing solvers lead to spurious results when the pressure levels decrease many orders of magnitude \cite{bellhop, jensen2011computational}. Such a rapid drop in pressure levels usually occurs at the boundary of the shadow zones where no ray traces enter. 
We observed spurious results as the pressure levels dropped beyond 10 orders of magnitude. Thus, pressure levels beyond 10 orders of magnitude were filtered away leading to a flat region in the transmission loss contour beyond 200 dB in Fig. \ref{fig:tl_soln_zs18}. For obtaining the data sets we considered Munk's sound speed profile sampled with 25 depth-wise velocities. Therefore, the ocean depth was divided into 24 depth-wise divisions with uniform fluid properties during BELLHOP's numerical ray trajectory computations.  Notably, the sound speed profile remains constant across the range, and these assumptions are deliberately chosen to examine the influence of range-dependent variations in bathymetry on the prediction of transmission loss. 

The underwater radiated noise depends on the geometry of ocean bathymetry. As the ocean profile changes, the transmission loss at a far-distance location varies dramatically. 
The presence of a seamount can significantly impact the resulting transmission loss in underwater acoustic environments. 
Seamounts can act as barriers, causing sound channels to form that can enhance sound propagation over long distances. 
Additionally, seamounts can create regions of quiet or reduced sound levels, particularly in their lee (downwind) sides, due to sound shadowing effects \cite{ebbeson1983sound}.
Understanding these effects is crucial for predicting and interpreting acoustic propagation in oceanic environments.
A neural network trained on underwater radiated noise data predicts the transmission loss not only for the training set but also for ocean bathymetry profiles outside of the training range (extrapolation). 
While a perfect extrapolation may be challenging in practice, we strive for a small extrapolation error. 
Figure \ref{fig:tl_soln_zs18} below shows the transmission loss estimated from the bellhop Gaussian beam solver for an ideal sea mount profile.


\subsection{Far-field transmission loss prediction for ideal seamount}

In this section, we evaluate the predictive accuracy of our range-dependent convolutional neural network framework across various ideal seamount bathymetry profiles. 
We examine three distinct test scenarios: Test Case 1 with peak depths at 450 m and 1650 m, Test Case 2 featuring varying base width with a peak depth of 1200 m and base located at [24 km, 43 km], and Test Case 3 representing a general seamount with a peak depth of 1600 m and base at [12 km, 44 km]. These test cases involve extrapolation beyond the bathymetry range covered in the training dataset.

\begin{figure}[h!]
    \centering
    \includegraphics[width=0.99\linewidth]{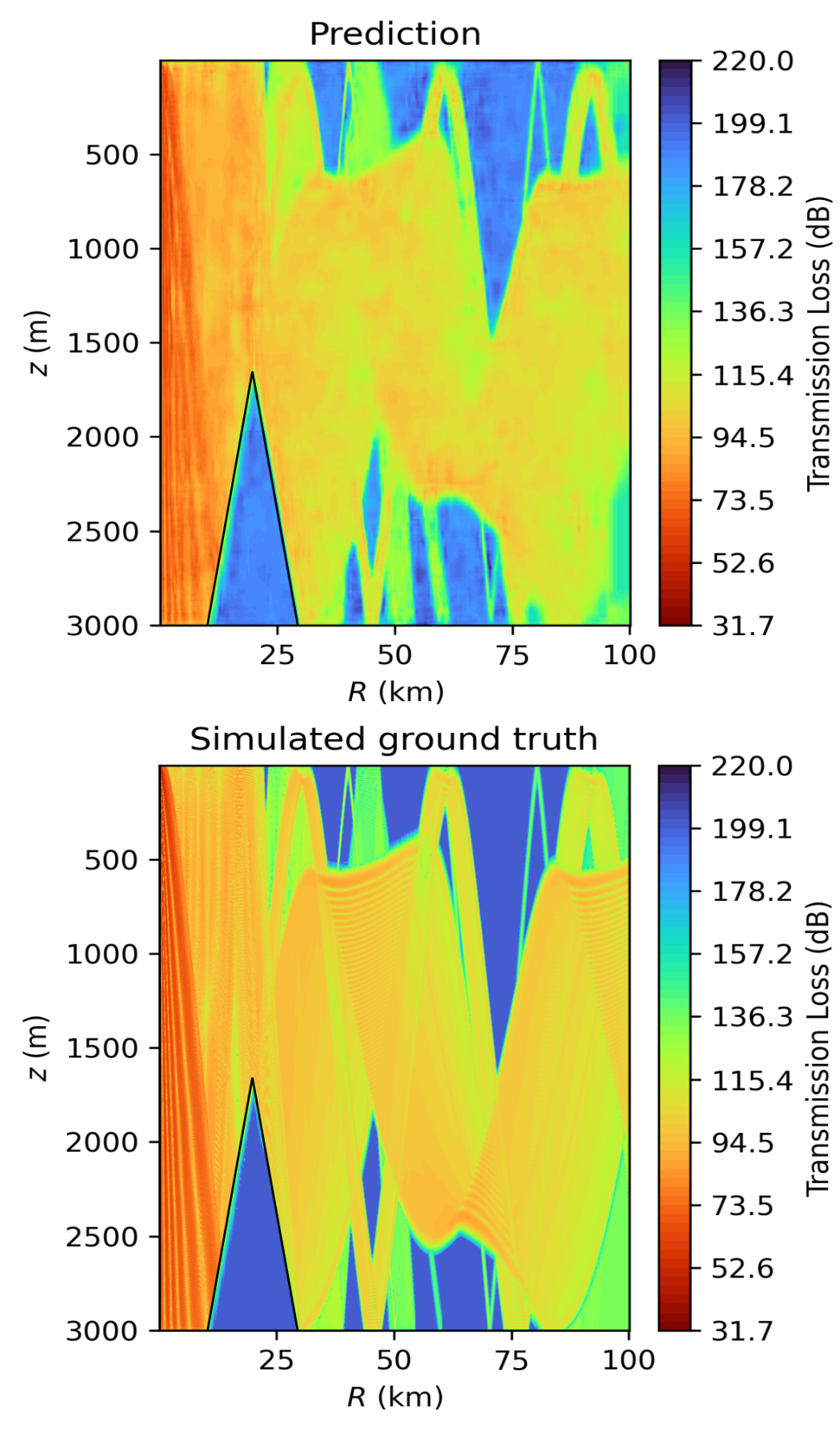}
    \caption{Peak depth 1650 m. Top, prediction from RC-CAN, and bottom, simulated ground truth from BELLHOP solver. }
\label{fig:SM_HV_T1}
\end{figure}

\begin{figure}[h!]
\centering
\subfigure[Probe depth = 500m]{\includegraphics[width=0.99\linewidth]{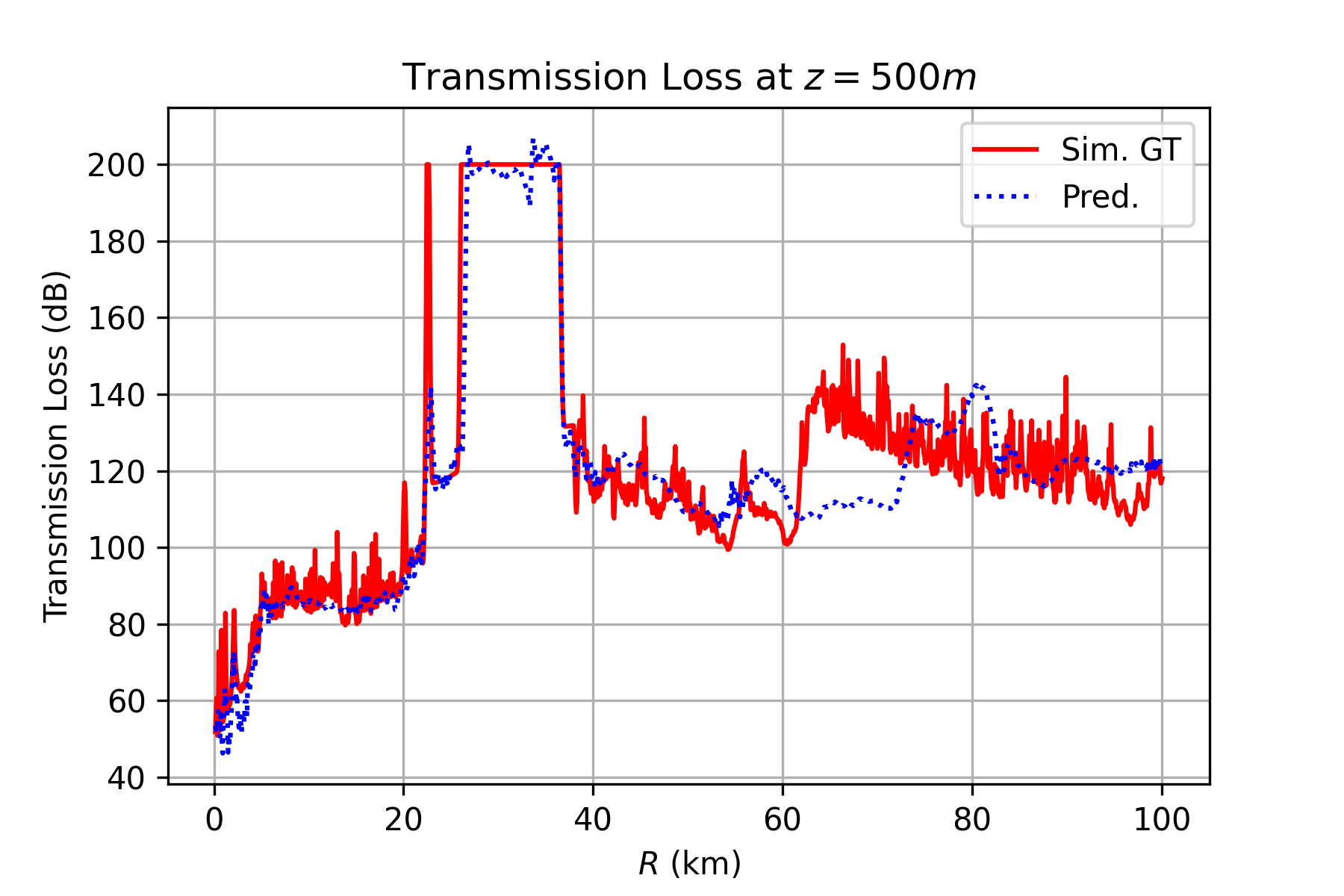}}
\subfigure[Probe depth = 1500m]{\includegraphics[width=0.99\linewidth]{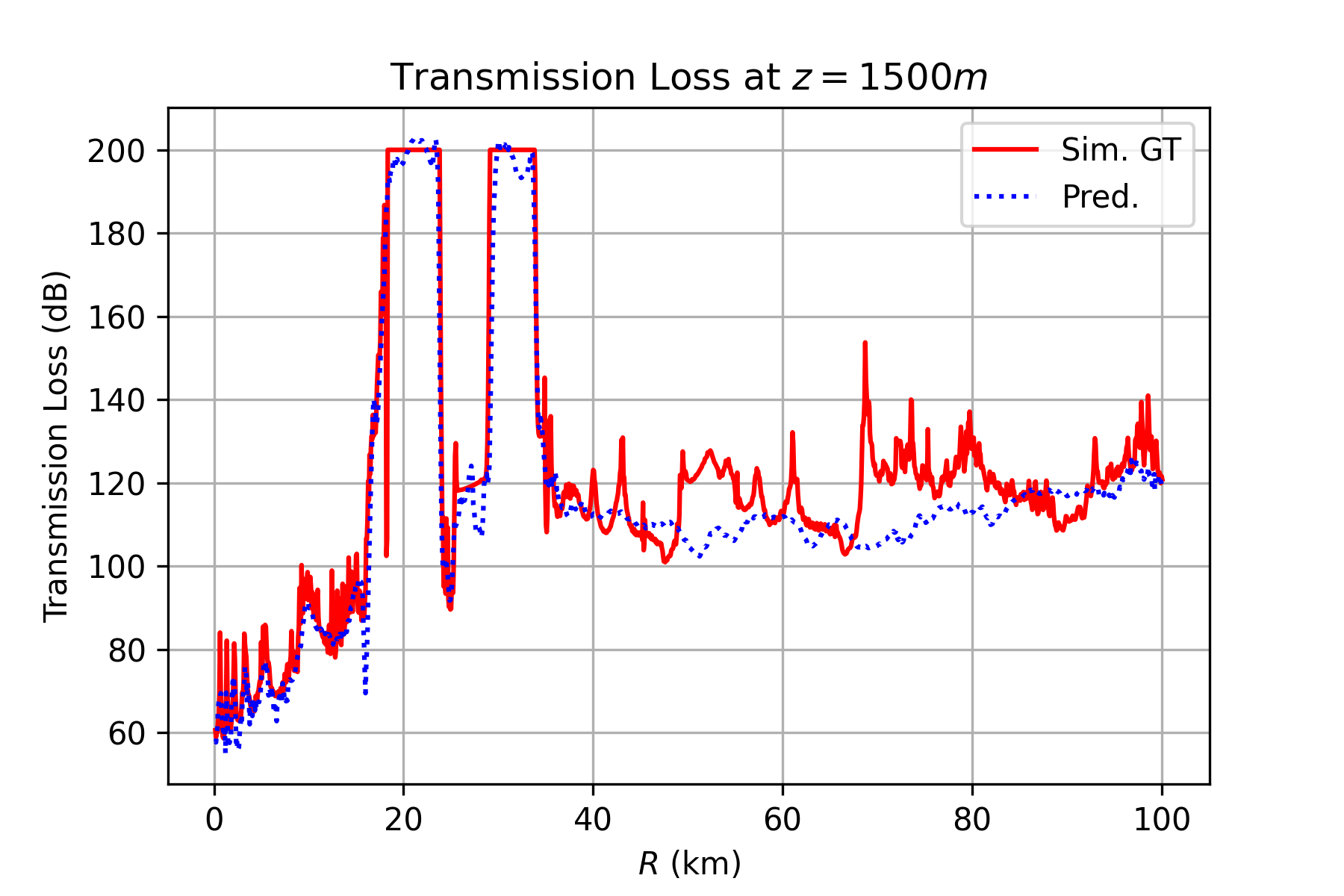}}

\caption{Transmission loss with range for peak depth of sea mount = 450 m. (a) Probe depth = 500m, and (b) probe depth = 1500m }
\label{fig:depth_500_1500}
\end{figure}

Starting with Test Case 1, characterized by a seamount's peak depth of 1650 m, Figure \ref{fig:SM_HV_T1} showcases a comparison between the simulated transmission loss and the predictions generated by the range-conditional convolutional neural network. Notably, our proposed architecture adeptly captures the underwater radiated noise propagation in this scenario, demonstrating its precision in modeling complex acoustic phenomena in the deep ocean environment. We measure the similarity between the prediction and simulation data using SSIM as mentioned previously. The SSIM in this case is 0.89 demonstrating that the prediction is 89 percent similar to the simulation data in the extrapolation regime. The red and warmer region in Figure \ref{fig:SM_HV_T1} corresponds to the least transmission loss and loudest noise region whereas the blue and cooler region represents maximum transmission loss and the quietest zone in the ocean domain. The RC-CAN as depicted in Figure \ref{fig:SM_HV_T1} correctly predicts the loud zones and the quiet zones which are important for building the adaptive noise mitigation strategy in real-time.
\newpage
In Test Case 1, where the peak depth is 450m, the structural similarity index (SSIM) value is 0.91, demonstrating the RC-CAN architecture's extrapolation capabilities. To delve deeper into the transmission loss characteristics, we proceeded to plot the transmission loss at two probe locations, one at a depth of 500m and another at 1500m.  Figure~\ref{fig:depth_500_1500} illustrates the variation of transmission loss with range at these probe depths. The plot indicates that RC-CAN effectively captures both the region of maximum loss and areas of low transmission loss. The convolution filter in the RC-CAN architecture introduces Gaussian smoothing, which filters out high-frequency content, as seen in Figure~\ref{fig:depth_500_1500}.  
As previously stated, there is an excellent level of agreement between RC-CAN predictions and simulated ground truth across the entire domain in both cases. However, there is a difference in the zone beginning at 60 km, where acoustic waves converge to form a loud zone. These loud zones indicate the convergence of acoustic rays, which contain high-frequency content that is usually smoothed by the convolution kernel. Despite this smoothing effect, RC-CAN can correctly identify the locations of these zones.

\begin{figure}[h!]
    \centering
    \includegraphics[width=0.99\linewidth]{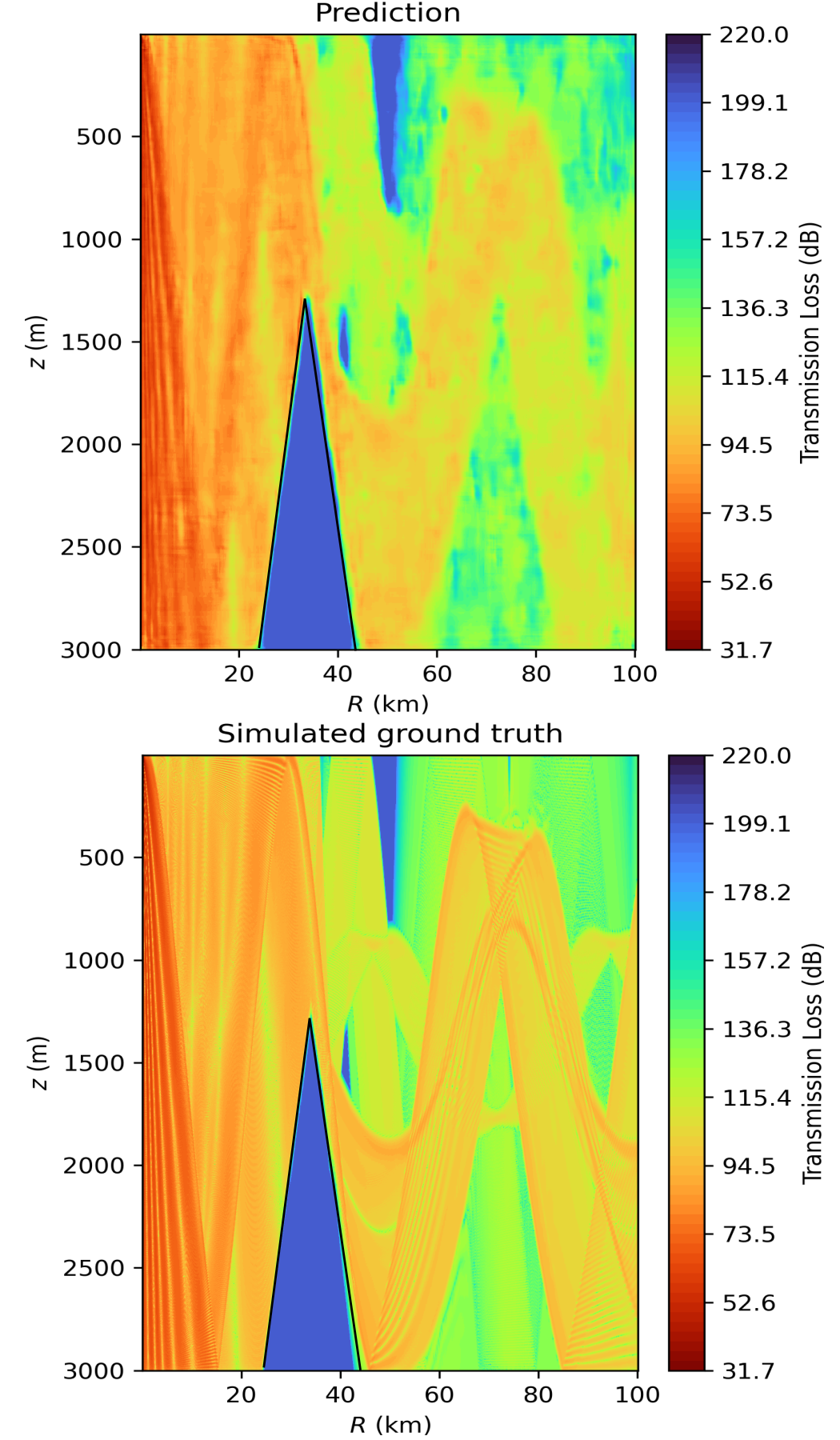}
    \caption{Peak's depth 1200 m and base at [24km, 43km]. Top, prediction from RC-CAN, and bottom, simulated ground truth from BELLHOP solver. }
\label{fig:SM_BV_T1}
\end{figure}

\begin{figure}[h!]
\centering
\subfigure[Probe depth = 500m]{\includegraphics[width=0.99\linewidth]{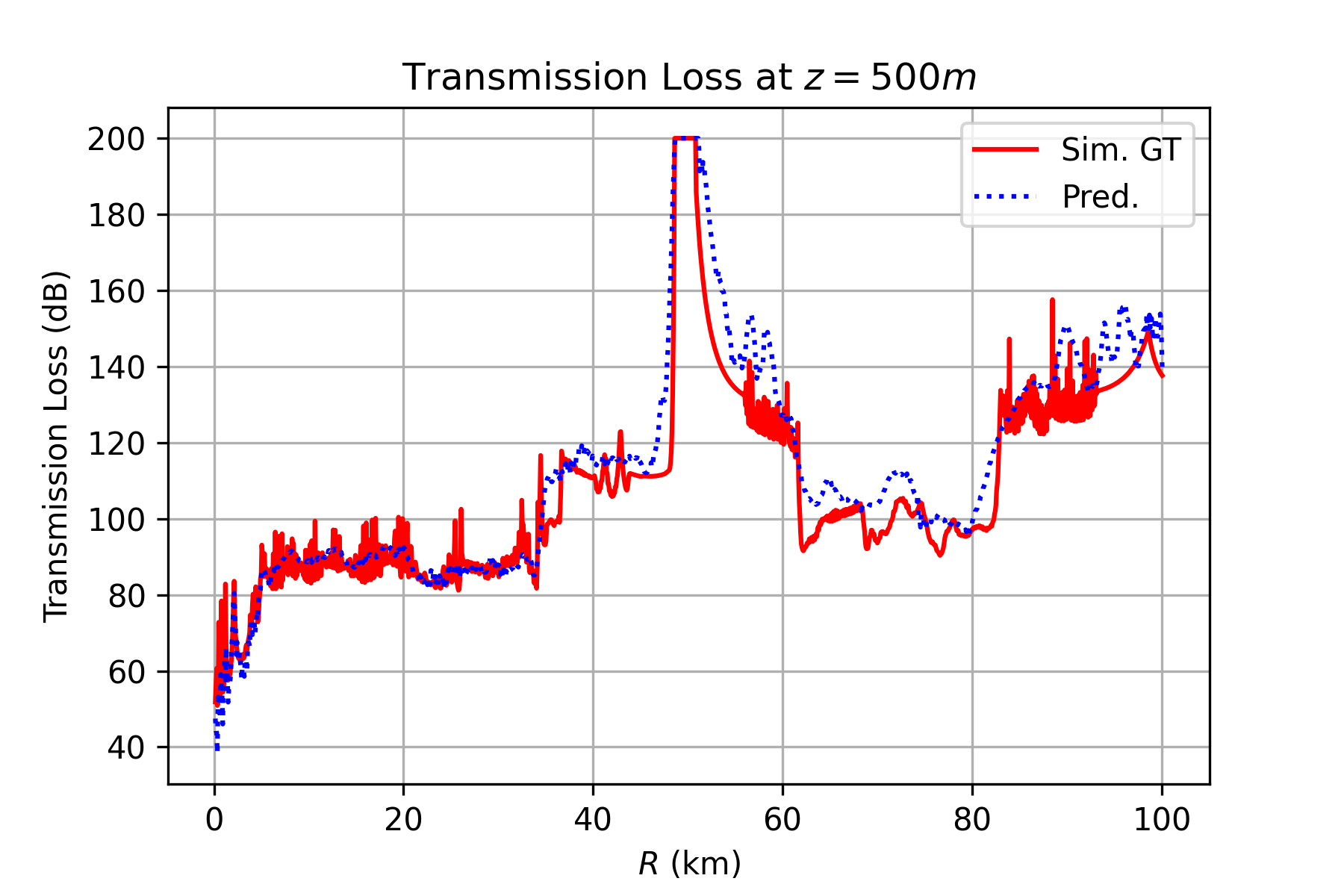}}
\subfigure[Probe depth = 1500m]{\includegraphics[width=0.99\linewidth]{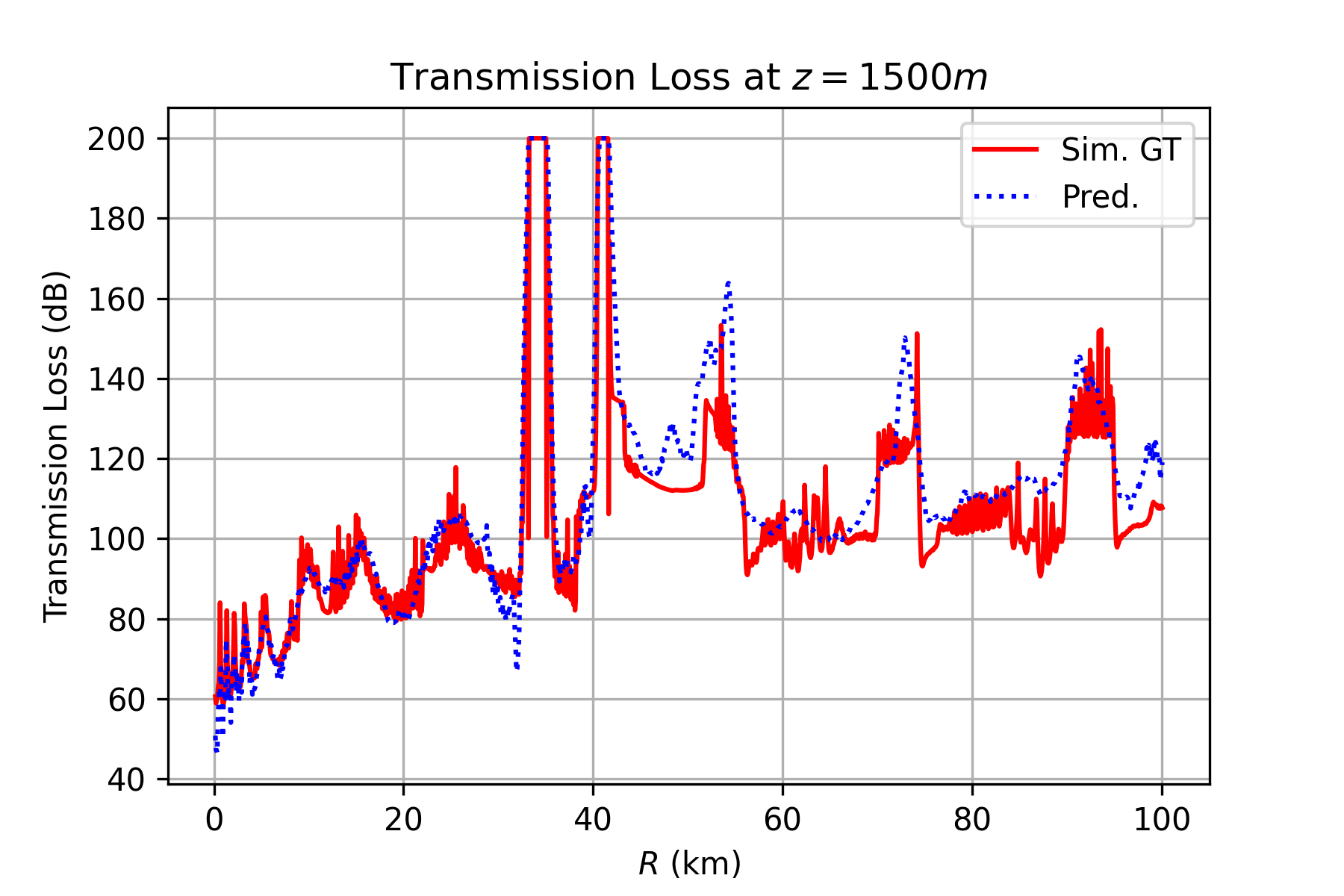}}

\caption{Transmission loss with range for peak depth of sea mount = 1200 m and base at [24km, 43km]. (a) Probe depth = 500m, and (b) probe depth = 1500m }
\label{fig:BV_depth_500_1500}
\end{figure}

For the test case where the base of the ideal seamount varies, with the base located at 24 km and 43 km and a peak depth of 1200m, the SSIM between RC-CAN and the simulation ground truth is 0.96. This result is depicted in Figure \ref{fig:SM_BV_T1}. Additionally, the transmission loss at depths of 500m and 1500m is analyzed and presented in Figure \ref{fig:BV_depth_500_1500} to gain further insights. Similar to the height-varying case, the variation of transmission loss with range is in good agreement with the simulation results from BELLHOP. This test case illustrates how the RC-CAN transmission loss prediction changes as the base of the seamount varies from the source of the sound.

\begin{figure}[h!]
\centering
\subfigure[Probe depth = 500m]{\includegraphics[width=0.99\linewidth]{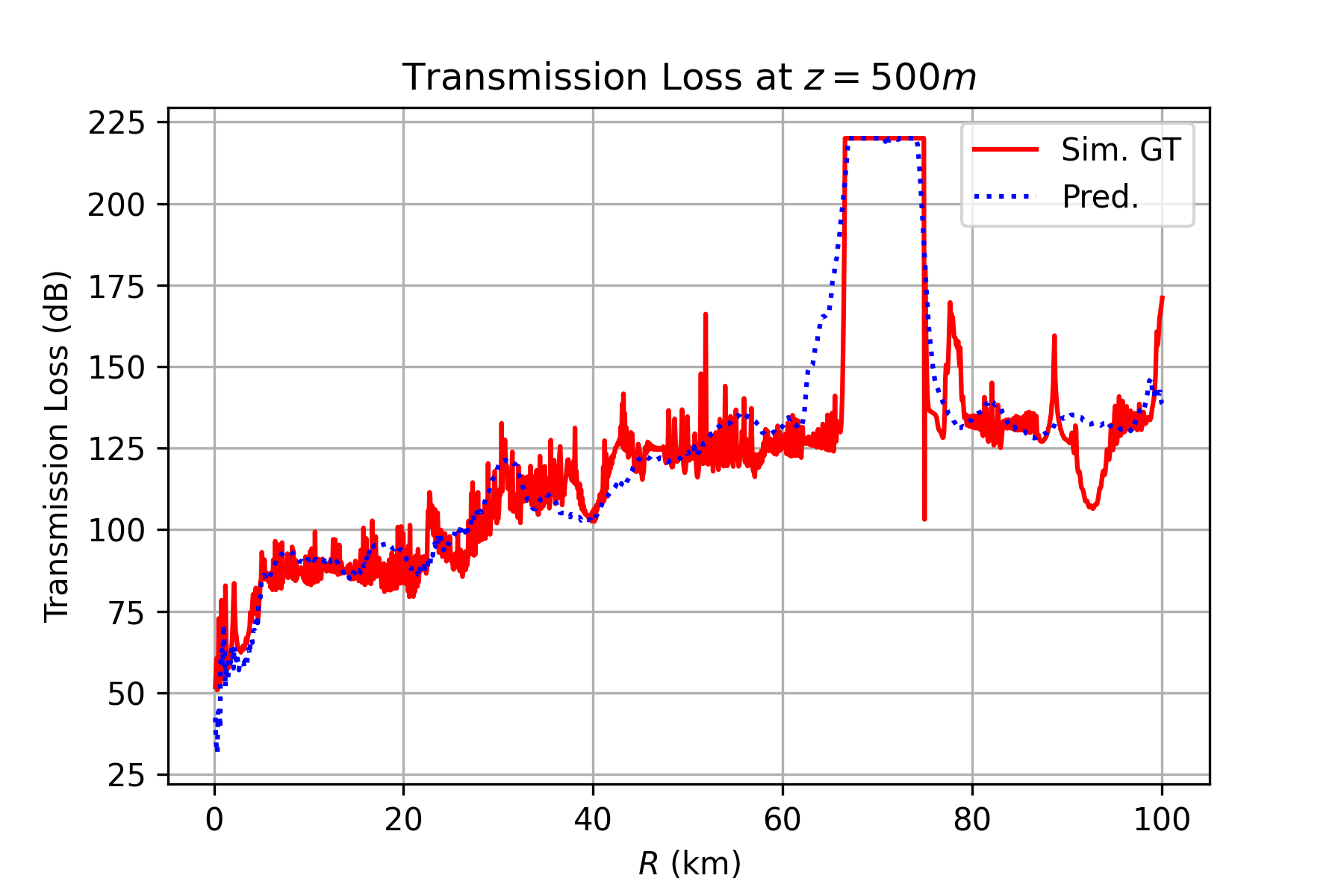}}

\caption{Transmission loss with range for peak depth of sea mount = 1600 m and base at [12km, 44km].}
\label{fig:Gen_depth_500}
\end{figure}

In Test Case 3, where both the peak depth and base of the seamount vary, the SSIM is 0.88. Figure \ref{fig:Gen_depth_500} illustrates the variation of transmission loss with range at a probe depth of 500m. There is a high level of agreement between the predictions of RC-CAN and the simulated ground truth across the entire domain. 
Across all test cases, the results underscore the superior predictive capabilities of the range-conditional convolutional neural network in capturing the complex dynamics of underwater noise propagation. The SSIM values calculated for all test cases are close to or greater than 0.9 in the extrapolation regime of bathymetry, demonstrating the model's ability to generalize across different seamount heights and bases. This positions it as a robust and advanced solution for accurate transmission loss prediction in diverse underwater environments. The effectiveness of our proposed architecture suggests its potential for predictive modeling in underwater acoustics, making it a valuable tool for marine engineering applications. 

\subsection{Far-field transmission loss prediction for wedge profile}
The ocean bathymetry features not only seamounts but also gradually sloping floors. A reliable model should be capable of accurately predicting sound propagation under these diverse bathymetric conditions. Unlike seamounts, the gradually sloping floor does not create shadows but rather represents areas where sound waves combine to form loud zones. Identifying these zones, especially during the movement of marine vessels, is crucial to minimize their impact on marine environments.

\begin{figure}[h!]
\centering
\includegraphics[width=0.99\linewidth]{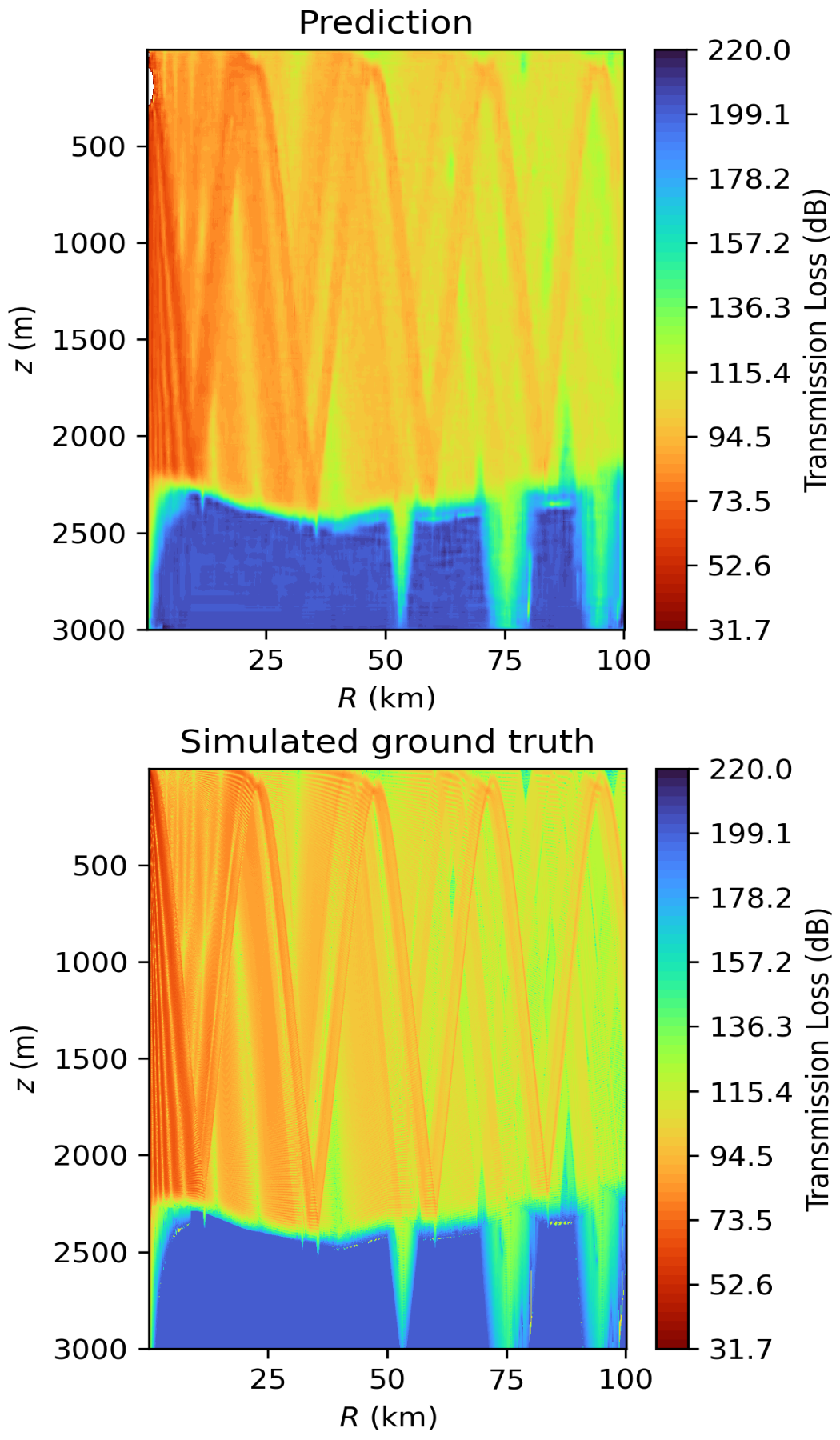}
\caption{Transmission loss prediction for test case 1. Top, prediction from RC-CAN, and bottom, simulated ground truth from BELLHOP solver.}
\label{fig:Wedge_Test1}
\end{figure}

To evaluate the model's performance in such scenarios, we designed a test case where the depth increases from 2180 m to 2380 m at 40 km and then gradually decreases back to 2100 m at a range of 100 km. The Structural Similarity Index (SSIM) between the prediction and RC-CAN for this case is 0.98, indicating that RC-CAN can accurately predict the trajectory of rays in this complex bathymetric scenario. Figure \ref{fig:Wedge_Test1} illustrates the transmission loss profile prediction by RC-CAN, showing a close alignment with the simulated ground truth.

\begin{figure}[h!]
\centering
\subfigure[probe depth = 500m ]{\includegraphics[width=0.45\textwidth]{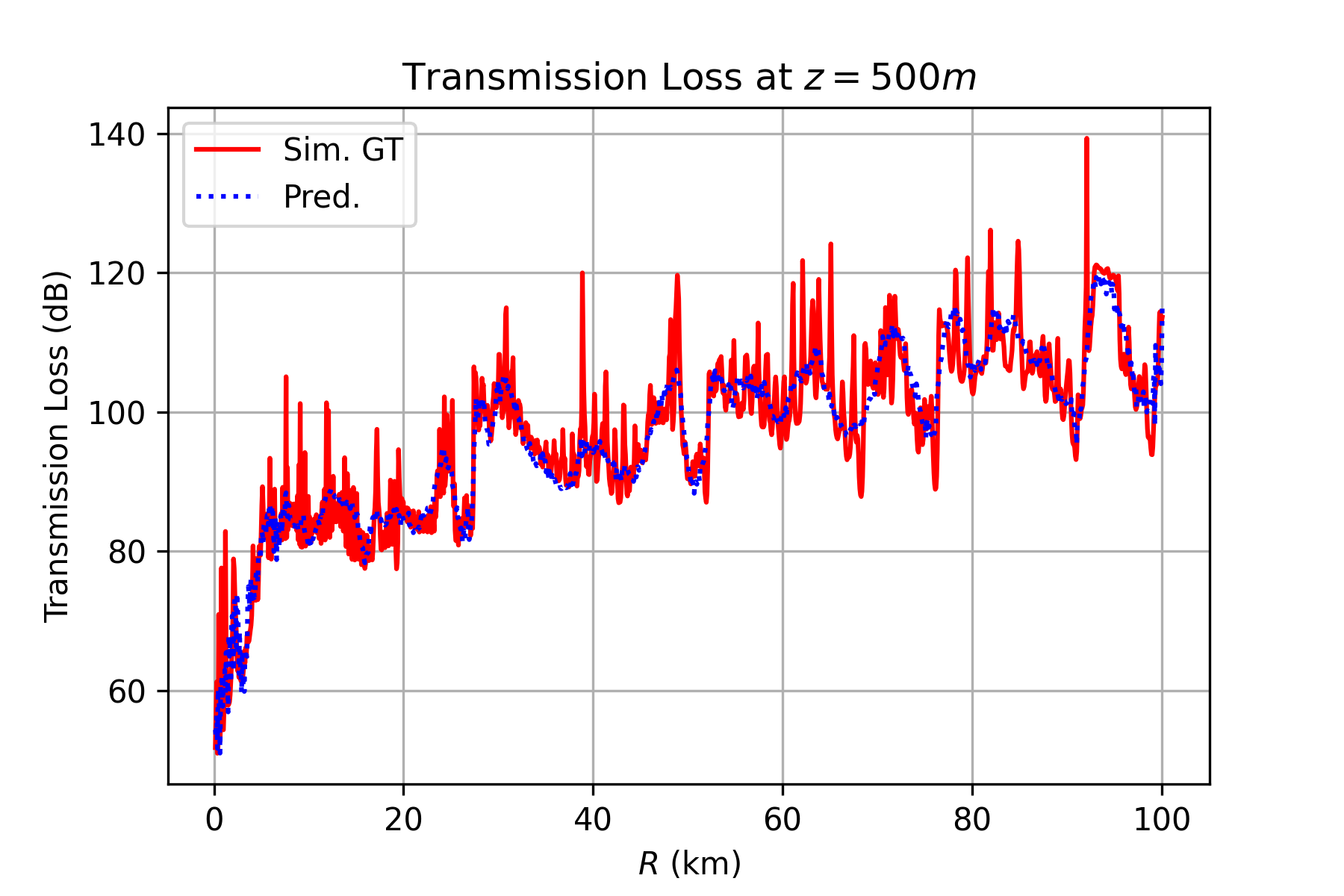}}
\subfigure[probe depth = 1500m ]{\includegraphics[width=0.45\textwidth]{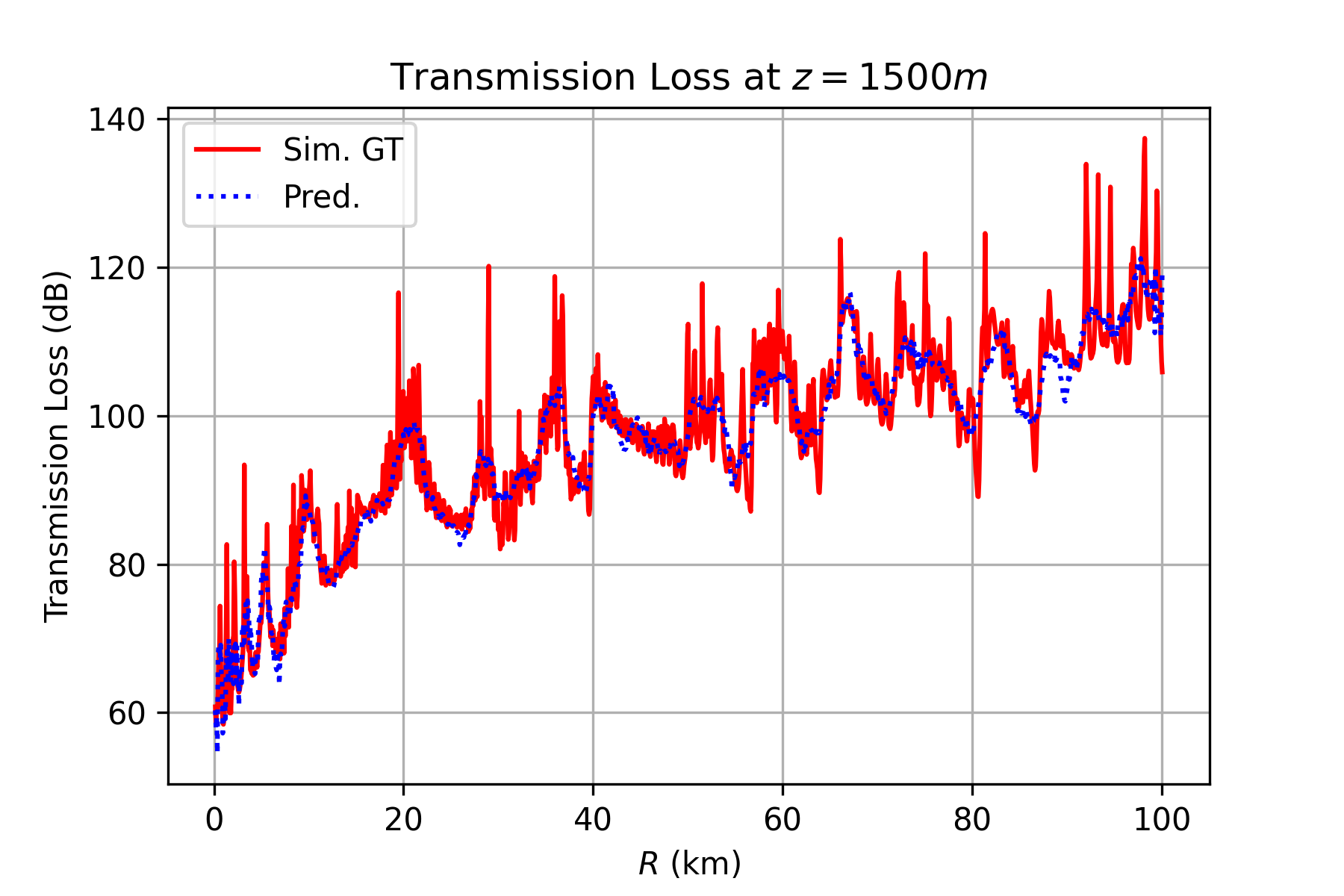}}
\caption{Transmission loss prediction for test case 1.}
\label{fig:Wedge_Test1_depth}
\end{figure}

Figure \ref{fig:Wedge_Test1_depth} further demonstrates the remarkable predictive performance of RC-CAN, closely aligning with the simulated ground truth across the entire domain. This test case highlights the model's capability to accurately predict sound propagation in scenarios involving gradually sloping floors, showcasing its robustness in handling diverse underwater environments.

\subsection{Dickins sea mount}
\begin{figure}[h!]
\centering
\includegraphics[width=0.99\linewidth]{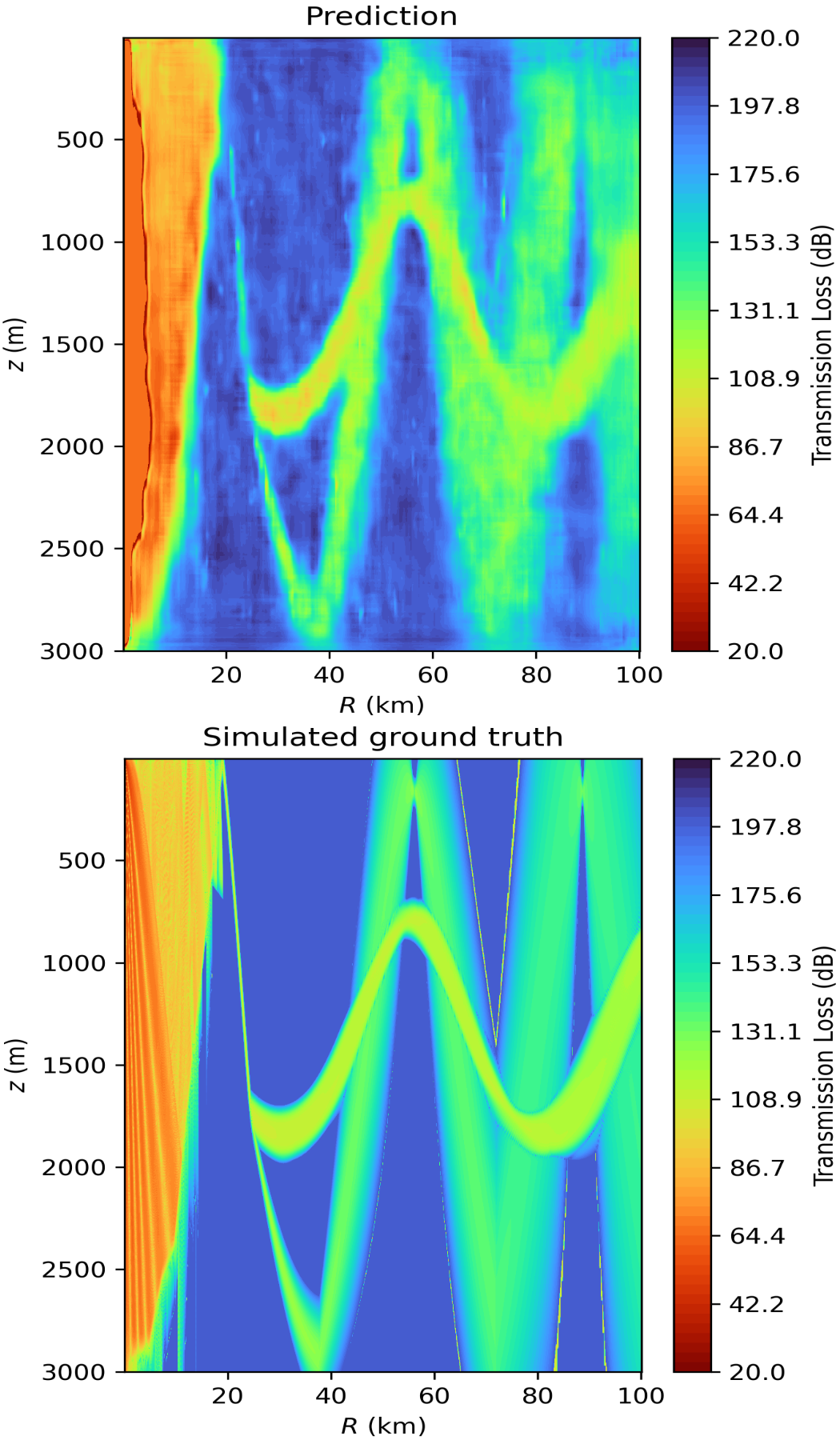}
\caption{Transmission loss prediction for Dickins sea mount. Top, prediction from RC-CAN, and bottom, simulated ground truth from BELLHOP solver.}
\label{fig:Dickins_pred}
\end{figure}

\begin{figure}[h!]
\centering
\subfigure[probe depth = 500m ]{\includegraphics[width=0.99\linewidth]{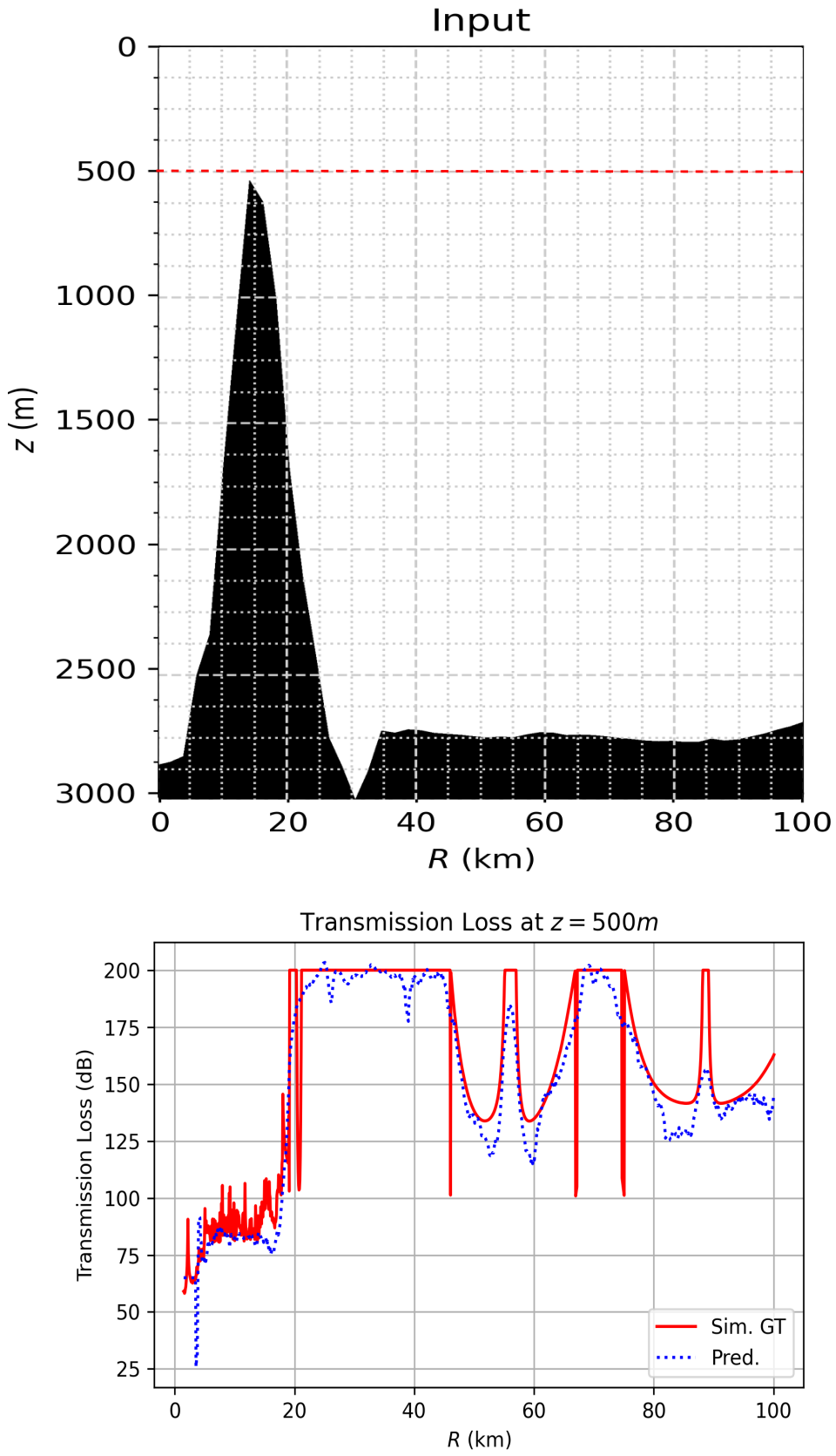}}
\caption{Transmission loss prediction for Dickins sea mount.}
\label{fig:Dickins_depth}
\end{figure}

It is crucial to have a model that accurately predicts the transmission loss profile for realistic ocean bathymetry. Seamounts are common features in the ocean. An example is Dickins Seamount, located in the Northeast Pacific Ocean \cite{ebbeson1983sound}.
To further evaluate the model's performance for realistic seamount profiles, we sampled the ocean bathymetry around the Dickins Seamount location. The Structural Similarity Index (SSIM) for the Dickins Seamount test case is 0.94, indicating good agreement with the simulation result. The prediction is illustrated in Figure \ref{fig:Dickins_pred}, while the transmission loss with range is depicted in Figure \ref{fig:Dickins_depth}. The RC-CAN model accurately identifies the location of ray convergence loud zones as well as the shadow zones.

The results from the Dickins Seamount test case highlight the robustness and reliability of the RC-CAN model in handling complex bathymetric features. Such capabilities are essential for accurately predicting transmission loss in real-world ocean environments, where seamounts and other underwater features can significantly impact sound propagation.


The mean SSIM accuracy for transmission loss in the test dataset is 90\%. Additionally, the RC-CAN model is able to compute the transmission loss distribution over the entire domain for each bathymetry within 0.01-0.5 CPU seconds. In comparison, similar predictions using BELLHOP required 75-150 CPU seconds for each bathymetry. Notably, the RC-CAN model is trained on the GPU and utilizes fast GPU computation for achieving real-time results. This significant difference in computation time highlights the potential application of the RC-CAN model for real-time decision-making and control in underwater acoustic environments.

\section{\label{sec:7} Conclusions}
In this article, the authors introduce the range-dependent conditional convolutional neural network (RC-CAN) for data-driven learning of far-field acoustic transmission in ocean environments with varying bathymetry. 
We propose a replay-based training strategy and range-dependent conditional convolutional neural network for modeling underwater radiated noise. The model effectively captures the complex relationship between ocean bathymetry and transmission loss. By transforming the input mesh into a latent space and decoding the transmission loss in a single step, the model demonstrates efficient learning capabilities.

When trained with full-domain snapshots of the transmission loss distribution, the RC-CAN model showed a generalized prediction capacity of far-field transmission loss for bathymetry outside the training set. Furthermore, such predictions were obtained in a physically consistent manner. Specifically, the RC-CAN model trained with transmission loss distribution obtained via ray/beam tracing solver Bellhop for idealistic bathymetry like ideal sea mount and gradually decreasing sea floor can predict transmission loss for Dickins sea mount with 90\% SSIM accuracy. 
Our empirical findings underscore the effectiveness of the single-shot learning strategy and the range-conditional architecture, both of which act as powerful implicit biases within the deep neural network. This results in enhanced generalization performance across varying ocean bathymetry conditions.
Throughout our study, the proposed model has consistently demonstrated efficient and accurate prediction of transmission loss.

We used medium-fidelity ray tracing solvers to get the ground truth for ocean acoustic transmission loss, which was then used to train and test the RC-CAN model. 
It's important to note that the RC-CAN model is data-driven and agnostic to how the data is acquired. Therefore, the RC-CAN model can seamlessly incorporate data obtained from higher-fidelity solvers or experimental measurements if available. This underscores the potential of data-driven deep learning models like RC-CAN to generate a digital twin of ocean acoustic transmission over a wide range of parameters and physical phenomena of varying complexity.
The real-time online prediction capability of such a digital twin holds promise for fast and physics-informed decision-making in marine vessel operations. By leveraging the RC-CAN model's ability to accurately predict transmission loss profiles in real-time, marine operators can make informed decisions to optimize vessel operations and minimize the impact on marine environments.

\begin{acknowledgments}
The present study is supported by Mitacs, Transport Canada, and Clear Seas through the Quiet-Vessel Initiative (QVI) program. We express our gratitude to Mr. Gaurav Bhatt from the UBC Department of Computer Science for his numerous invaluable suggestions and discussions regarding the framework for continual learning. The authors would like to express their gratitude to Dr. Paul Blomerous and Ms. Tessa Coulthard for their valuable feedback and suggestions. Additionally, we gratefully acknowledge the contributions of Dr. David Rosen and Dr. Andrew Trites, who provided insights on the marine biology and conservation aspects of the study. We would also like to acknowledge that the GPU facilities at the Compute Canada clusters were used for the training of our deep learning models.
\end{acknowledgments}

\section*{References}
\bibliography{sampbib}

\end{document}